\documentclass[letterpaper, 10 pt, conference]{IEEEtran} 
\usepackage{booktabs}
\usepackage{multirow}
\usepackage{siunitx}
\usepackage{caption} % Add this package
\usepackage{etoolbox}

\usepackage{soul}
\usepackage{xcolor}
\usepackage{cancel} 
\usepackage[normalem]{ulem} 

\newcommand{\removecomment}[1]{}
\newcommand{\editcomment}[2]{{#1}}

\newcommand{\citationremove}[1]{}

\usepackage[font=small,skip=2pt]{caption}

\IEEEoverridecommandlockouts                              % This command is only needed if 
\usepackage[T1]{fontenc}
\usepackage{graphics}
\usepackage{amsmath}
\usepackage{amsfonts}
\usepackage{colortbl}
\usepackage{amssymb} % to use triangleq
\usepackage{graphicx}
\usepackage{enumerate}   
\makeatletter
\let\NAT@parse\undefined
\makeatother
\usepackage{hyperref}
\usepackage[sort,compress]{cite}

\usepackage{lipsum}

\usepackage{balance}
\usepackage{arydshln}

\usepackage{booktabs}
\usepackage{tikz}

\usepackage{caption}
\usepackage{mathrsfs}

\usepackage[captionskip=2pt]{floatrow}
\newlength\Myfigwd
\usepackage{tabularx}
\newcommand{\realfield}{\hbox{I \kern -.4em R}}
\newcommand {\mb}[1]{\mathbf{#1}} % all replaced
\newcommand {\bs}[1]{\boldsymbol{#1}}
\newcommand{\uvec}[1]{\hat{\mathbf{#1}}}

\newcommand{\T}{^{\top}}  %shortcut for transpose
\newcommand*{\diameter}{\bigcirc\kern-0.95em\diagup}
\newcommand{\rmd}{\textrm{d}}  %shortcut for derivative
\makeatletter
\newcommand{\rom}[1]{\uppercase\expandafter{\romannumeral #1\relax}}
\makeatother
\usepackage{algorithm}
\usepackage{algpseudocode}

\makeatletter
\newcommand*{\algrule}[1][\algorithmicindent]{\makebox[#1][l]{\hspace*{.5em}\vrule height .75\baselineskip depth .25\baselineskip}}%

\newcount\ALG@printindent@tempcnta
\def\ALG@printindent{%
    \ifnum \theALG@nested>0% is there anything to print
        \ifx\ALG@text\ALG@x@notext% is this an end group without any text?
            \addvspace{-3pt}% FUDGE for cases where no text is shown, to make the rules line up
        \else
            \unskip
            \ALG@printindent@tempcnta=1
            \loop
                \algrule[\csname ALG@ind@\the\ALG@printindent@tempcnta\endcsname]%
                \advance \ALG@printindent@tempcnta 1
            \ifnum \ALG@printindent@tempcnta<\numexpr\theALG@nested+1\relax% can't do <=, so add one to RHS and use < instead
            \repeat
        \fi
    \fi
    }%
\usepackage{etoolbox}
\patchcmd{\ALG@doentity}{\noindent\hskip\ALG@tlm}{\ALG@printindent}{}{\errmessage{failed to patch}}
\makeatother
\usepackage{multirow}  %needed for multi-row table
\usepackage{tabularx}  %needed for tabularx environment for nicer column alignment in a table
\usepackage{colortbl} %needed for colored cells in Tabularx see http://tex.stackexchange.com/questions/148784/colored-diagonal-divided-table-cell
\usepackage{footnote}% placed after tabularx so it does not conflict
\makeatletter
\newcommand{\thickhline}[1]{%
    \noalign {\ifnum 0=`}\fi \hrule height #1
    \futurelet \reserved@a \@xhline
}
\newcolumntype{"}{@{\hskip\tabcolsep\vrule width 2pt\hskip\tabcolsep}}
\makeatother 
\graphicspath{{figures/pdf_files/}}
\begin{document}
%
% paper title
% Titles are generally capitalized except for words such as a, an, and, as,
% at, but, by, for, in, nor, of, on, or, the, to and up, which are usually
% not capitalized unless they are the first or last word of the title.
% Linebreaks \\ can be used within to get better formatting as desired.
% Do not put math or special symbols in the title.
\title{Stochastic Adaptive Estimation in Polynomial Curvature Shape State Space for Continuum Robots}%
%
% author names and IEEE memberships
% note positions of commas and nonbreaking spaces ( ~ ) LaTeX will not break
% a structure at a ~ so this keeps an author's name from being broken across
% two lines.
% use \thanks{} to gain access to the first footnote area
% a separate \thanks must be used for each paragraph as LaTeX2e's \thanks
% was not built to handle multiple paragraphs
%

\author{Guoqing~Zhang and Long~Wang% <-this % stops a space
\thanks{G. Zhang and L. Wang are with the Department
of Mechanical Engineering, Stevens Institute of Technology, New Jersey,
NJ, 07030, USA, e-mail: lwang4@stevens.edu}% <-this % stops a space
\thanks{This work was supported in part by USDA-NIFA Grant No.~2021-67022-35977, NSF Grant CMMI-2138896, and Stevens Institute of Technology internal grants.}
}

\maketitle
% \tableofcontents
\begin{abstract}
% Sentence 1
In continuum robotics, real-time robust shape estimation is crucial for planning and control tasks that involve physical manipulation in complex environments. 
% Sentence 2
In this paper, we present a novel stochastic observer-based shape estimation framework designed specifically for continuum robots. 
% Sentence 3
The shape state space is uniquely represented by the modal coefficients of a polynomial, enabled by leveraging polynomial curvature kinematics (PCK) to describe the curvature distribution along the arclength. 
% Sentence 4
Our framework processes noisy measurements from limited discrete position, orientation, or pose sensors to estimate the shape state robustly. 
% Sentence 5
We derive a novel noise-weighted observability matrix, providing a detailed assessment of observability variations under diverse sensor configurations. 
% Sentence 6
To overcome the limitations of a single model, our observer employs the Interacting Multiple Model (IMM) method, coupled with Extended Kalman Filters (EKFs), to mix polynomial curvature models of different orders. 
% Sentence 7
The IMM approach, rooted in Markov processes, effectively manages multiple model scenarios by dynamically adapting to different polynomial orders based on real-time model probabilities. 
% Sentence 8
This adaptability is key to ensuring robust shape estimation of the robot's behaviors under various conditions. 
% Sentence 9
Our comprehensive analysis, supported by both simulation studies and experimental validations, confirms the robustness and accuracy of our methods.

\end{abstract}
\section{Introduction}
\par Inspired by biological features such as an elephant's trunk or an octopus's tentacles, continuum robots are an unconventional class of robots that have inherent large degrees of flexibility, underactuation, and compliance. Such "soft" features allow them to have safe physical interactions with the environments\cite{hirose1993biologically}. With recent advances in instrument miniaturization, continuum robot designs have been widely used in many minimally invasive surgery applications \cite{simaan2018medical,dupont2022continuum}, and their advantages are further leveraged for safe manipulation and access to deep anatomy. \par
Robustly ascertaining the shape of continuum robots is critical for effective motion planning and control.
This task typically involves a comprehensive shape estimation framework integrating sensing techniques, representation methods, and estimation approaches.
\subsection{Sensing Technique}
\par Shape sensing is pivotal in this context by supplying the observational data critical for robust shape estimation. A diverse array of sensing techniques is deployed to acquire the shape information. External sensing involves equipping robots with sensors such as Electromagnetic (EM) sensors\cite{song2015electromagnetic,anderson2017continuum,lilge2022continuum,borgstadt2015multi,ataka2016real,shi2016shape}, vision-based systems\cite{venkiteswaran2019shape,borgstadt2015multi,loo2019non}, or Inertial Measurement Units (IMUs)\cite{cheng2022orientation,peng2024tendon} to obtain pose information externally. Alternatively, intrinsic sensing leverages internal sensors to enhance joint-level information, utilizing encoders or passive cable displacement measurements \cite{orekhov2023lie}.
Given their infinite degrees of freedom (DoF), the challenge with continuum robots lies in capturing comprehensive pose information along their entire length with a finite number of sensors. While technologies like Fiber Bragg Grating (FBG) sensors\cite{kim2014optimizing,qiao2019estimating,qiao2021force,al2021fbg,sefati2017highly,shi2016shape} and vision-based systems can map out the full robot shape, constraints such as budget and the specific requirements of the application often preclude their ubiquitous use. \par
In this work, we explicitly target limited-sensing deployments, using sparse pose/position/orientation measurements (in our experiments, a single tip measurement) to reflect these practical constraints.
\subsection{Representations in Modeling}
Different shape representation methods yield various formulations of the backbone curvature, which directly impact continuum robots' computed position and orientation.
Representation methods can be categorized as distributed and lumped parameterizations for the backbone curvature\cite{rao2021model, rone2013continuum}. \par
Distributed parameterization methods represent the backbone curvature as a continuous function of the arc length, characterized by an infinite number of parameters. These methods assume that the curvature distribution is variable along the length of the robot.
A classic example is the Cosserat rod theory (CRT), which employs distributed parameter modeling to account for the underlying mechanics by solving for positional, force, and moment boundary conditions \cite{jones2009three,till2019real,boyer2020dynamics,renda2018discrete,rucker2011statics}. \par
Lumped parameterization uses a finite set of parameters to describe the backbone shape of continuum robots geometrically. This approach primarily includes three methods: the Pseudo-Rigid-Body Model (PRBM), the constant curvature (CC) assumption, and the modal approach\cite{rao2021model}.
PRBM approximates the central backbone shape of continuum robots as a series of rigid links connected by torsion springs\cite{huang20193d,su2009pseudorigid,khoshnam2015robotics,roesthuis2016steering}.
The CC model assumes that each segment of the continuum robot follows a constant curvature, typically represented as a circular arc, with a single curvature parameter defining the shape.
Due to its simplification and computation efficiency, it is deployed quickly for ideal situations such as uniform mass/material distribution, frictionless transmission, and free from external loading. %
As a widely adopted kinematics-based approach, The CC model is extensively used in applications ranging from surgical procedures \cite{simaan2004dexterous,sears2006steerable,camarillo2008mechanics,webster2010design} and field operations\cite{hannan2003kinematics} to underwater exploration\cite{sitler2022modular} and aerial maneuvers \cite{zhao2022modular}. 
In contrast, the modal approach\cite{chirikjian1994modal} represents backbone curvature as a linear combination of modal shape functions (e.g., monomials\cite{wang2019geometric}, Euler curves \cite{rao2022shape}, Chebyshev polynomials\cite{orekhov2023lie}, and polynomial curvature kinematics (PCK)\cite{cheng2022orientation, della2019control}).
A modal approach offers the advantage of capturing complex, continuous deformations with a relatively small number of parameters, making it computationally efficient while still maintaining high accuracy in shape representation.
 \par
In this work we adopt the polynomial-curvature (PCK) modal family as the lumped parameterization; our experiments bring prior PCK theory into hardware validation under limited/sparse sensing, responding to calls for experimental studies and fair comparisons among polynomial-shape methods~\cite{della2019control,sadati2023reduced}.
\subsection{Estimation Approach}
Various sensing techniques and representation methods drive multiple estimation approaches tailored to the specific scenario. 
In static configurations, where the continuum robot remains stationary and its state does not change over time, researchers often employ curve fitting or optimization methods based on modal approaches\cite{kim2014optimizing, song2015electromagnetic, rao2022shape, orekhov2023lie}, and PRBM\cite{venkiteswaran2019shape}.
Considering model uncertainty and measurement noise, Lilge et al. \cite{lilge2022continuum} and Ferguson et al. \cite{ferguson2024unified} implemented Gaussian Process Regression (GPR) for static shape estimation under arclength domain using CRT. \par
Dynamic configurations involve continuous state updates as the continuum robot moves and its state changes over time.
In such scenarios, stochastic observer-based estimation methods, particularly filtering approaches, are preferred for their ability to optimize state estimation by balancing the influences of process and measurement models, where noise from both sources significantly impacts accuracy and robustness.
Although the Extended Kalman Filter (EKF) is a general estimation technique that can be paired with various forward models, most existing continuum robot shape-estimation studies have employed EKF in conjunction with CC or piecewise constant curvature (PCC) models\cite{ataka2016real, loo2019non, peng2024tendon}. In addition to the EKF, alternative filtering methods—such as the Unscented Kalman Filter (UKF) and particle filters—have also been explored \cite{chen2019model, borgstadt2015multi}. Each technique presents trade-offs in computational complexity, robustness, and ease of implementation. For instance, UKF’s sigma-point method can improve estimation in highly nonlinear systems, whereas particle filters can capture non-Gaussian noise at the expense of increased computation time. In certain cases, the EKF remains a better choice when the system model is sufficiently accurate, or when computational resources are limited.
Another well-established multi-model estimation approach is the Interacting Multiple Model (IMM) approach \cite{mazor1998interacting, dingler2022state}. Initially developed for tracking targets with Markovian switching dynamics, IMM simultaneously executes multiple Kalman-type filters (e.g., EKF or UKF), each associated with distinct motion models. Estimates from these parallel filters are integrated based on model probabilities updated via measurement innovations, commonly determined through likelihood or Normalized Innovation Squared (NIS) tests. IMM-based approaches are advantageous when a system switches between distinct kinematic regimes or experiences abrupt behavioral changes, conditions frequently encountered in continuum robotics due to varying loads, external interactions, or segmental bending transitions. Despite extensive IMM applications in mobile robotics and trajectory tracking, its deployment in dynamic shape estimation for continuum robots remains scarce, highlighting a valuable research direction in robust multi-model estimation.\par
In this paper, we treat shape estimation as a multi-model problem in curvature space, running an IMM-EKF over the PCK family (e.g., CC, PCK-1, PCK-2). For clarity, we compare against CC EKF and single-PCK EKFs using the same PCK models (without IMM) to isolate the benefit of multi-model mixing.

\subsection{Key Gaps in Shape Estimation Methods}
Several critical gaps persist in current continuum robot shape estimation:\par
Gap 1 lies in the lack of a curvature-space stochastic estimation framework tailored to limited sensing.
Most existing approaches rely on constant-curvature (CC) or piecewise-constant-curvature (PCC) models, focusing on pose in Cartesian or joint space. These methods do not fully capture a continuum robot’s shape in curvature space, which more directly reflects the robot’s natural mechanics. A dedicated stochastic estimation framework in curvature space remains missing, limiting accuracy and robustness in challenging scenarios.

Gap 2 lies in the limited use of multiple-model Kalman filtering for continuum robots. Although multiple-model Kalman filtering (e.g., IMM) is well established in other robotics contexts, it has not been systematically applied to continuum robots. Parallel filtering strategies could address the nonlinearities and model variations inherent in curvature-based representations, yet this approach remains underexplored in current research.

Gap 3 lies in the absence of rigorous curvature-space observability analysis for sensor optimization. 
Current observability methods commonly operate in Cartesian space and may overlook noise propagation specific to curvature-based models. A noise-weighted, curvature-space observability matrix can guide sensor configuration under sparse measurements.
\subsection{Contributions of This Work}
To address the challenges in dynamic shape estimation across various types of continuum robots, we propose a pioneering stochastic observer-based framework that utilizes general pose, position, and orientation sensing techniques combined with PCK. The principal contributions of our research include:

\begin{itemize}
\item Stochastic observer in polynomial curvature shape space. We directly estimate the modal (polynomial) coefficients of a continuum robot’s curvature in limited-sensing settings. By allowing higher-order polynomial terms, the method captures complex bending more accurately than constant-curvature models and remains robust to process and measurement uncertainties.

\item Adaptive multi-model estimation (IMM-EKF). We pose dynamic shape estimation as a multi-model problem in curvature space and run an IMM-EKF over the PCK family (CC/PCK-1/PCK-2) to adapt across regimes. We report baselines with CC EKF and single-PCK EKFs using the same PCK models (without IMM), isolating the benefit of multi-model mixing. To the best of our knowledge, this is the first application of IMM to curvature-space shape estimation for continuum robots.

\item Noise-weighted curvature-space observability. We derive a curvature-space observability matrix that incorporates sensor noise to inform sensor configuration under sparse measurements.
\end{itemize}
\par

\subsection{Scope and Applicability}
\textit{Scope.} We address shape estimation under limited/sparse sensing (pose/orientation/position) in a predominantly planar-bending regime. Out-of-plane curvature and twist are not explicitly estimated and are treated as process disturbances; experiments were designed accordingly. Claims are confined to this operating envelope.

\textit{Applicability.} The IMM–EKF in curvature space is platform-agnostic. Porting to other continuum robots (e.g., pneumatic, concentric-tube) requires updating platform kinematics and measurement Jacobians; the estimator core (polynomial-curvature state, process/measurement updates, IMM model mixing) and use of sparse measurements remain unchanged.

\section{Problem Formulation}
To develop a robust stochastic observer-based shape estimation framework for continuum robots, enhancing their functionality in dynamic and unpredictable environments, four specific problems need to be addressed, which are framed around the following foundational equations:
\begin{align}
& \dot{\mb{m}}(t)=\mb{f}(\mb{m}(t), \mb{u}(t)) 
\label{eqn:problem_formulation_process}\\
& \mb{y}(t)=\mb{h}(\mb{m}(t))
\label{eqn:problem_formulation_measurement}\\
& \hat{\mb{m}}(t)=\text{Observer\_Method}(\cdot)
\label{eqn:problem_formulation_observer}
\end{align}
where
\begin{itemize}
   \item[-] $t$ represents time,
   \item[-] $\mb{m}(t)$ is the robot's shape state vector, containing variables that describe the shape of the robot,
   \item[-] $\mb{u}(t)$ denotes the system input,
   \item[-] $\mb{y}(t)$ is the output vector, which can be directly measured from the system,
   \item[-] $\dot{\mb{m}}(t)$ denotes the derivative of $\mb{m}(t)$ with respect to time, describing the system's dynamics,
   \item[-] $\mb{f}$ is a vector function defining the state dynamics, depending on the current state and input,
   \item[-] $\mb{h}$ is a measurement function mapping the state space to the measurement space,
   \item[-] $\hat{\mb{m}}(t)$ is the estimate of the shape state,
   \item[-] $\text{Observer\_Method}(\cdot)$ denotes the observer algorithm employed to estimate the state vector from the measurements and the process model.
   
\end{itemize}
These problems include:
\begin{enumerate}
    \item \textit{Shape state representation}: Defining a compact and computationally efficient shape state vector $\mb{m}(t)$ that can accurately capture the dynamic properties of the continuum robot. PCK is utilized to address this problem (details are provided in Section \ref{ch:pc_kiematics}).
    \item \textit{Process model development}: Constructing a process model that accurately describes the dynamics of the robot's shape changes over time, influenced by control inputs $\mb{u}(t)$ and inherent system noise. This model builds upon the state transition equation (\ref{eqn:problem_formulation_process})(the detailed derivation is available in Section \ref{ch:state_transition})  with process noises.
    \item \textit{Measurement model construction}: Creating a measurement model that maps the shape state to observable outputs, ensuring that this model accounts for potential measurement noise. This model is constructed based on the output equation (\ref{eqn:problem_formulation_measurement})(the detailed derivation is presented in Section \ref{ch:measurement}) with sensor noises.
    \item \textit{Stochastic observer design}: architecting an adaptive recursive estimation approach (equation \ref{eqn:problem_formulation_observer}) to accurately estimate the shape state based on the process and measurement models. Interacting multiple models (IMM) coupled with EKFs are employed to address this challenge (detailed discussions are illustrated in Section \ref{ch:recursive_est_desgin}).
\end{enumerate}
\par
\begin{figure}[!t]
	\centering
	\includegraphics[width=0.9\columnwidth]{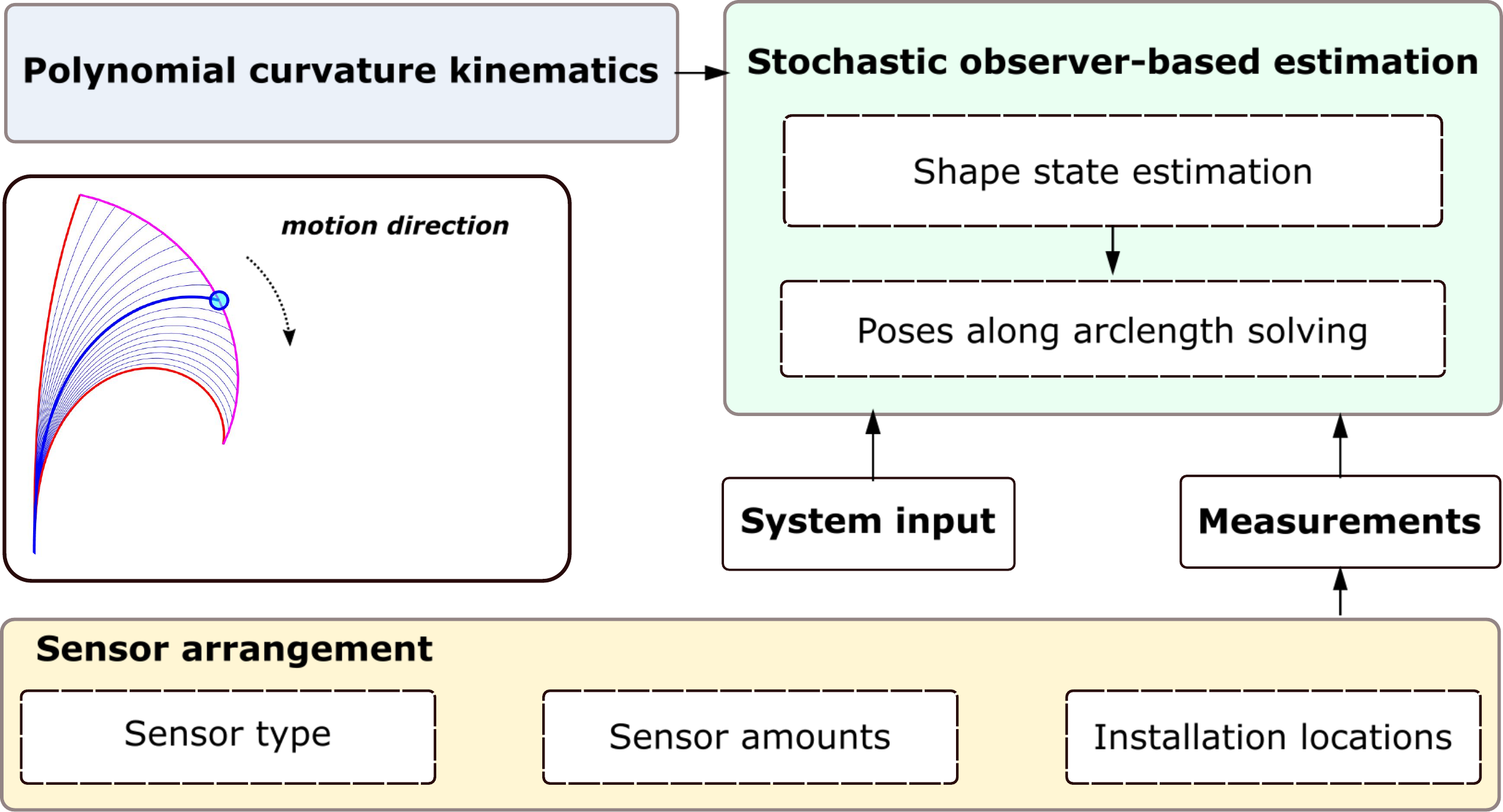}
	\caption{Diagram of proposed shape estimation method }
	\label{fig:shape estimation diagram}
\end{figure}

The detailed workflow of the shape estimation methodology is depicted in Fig. \ref{fig:shape estimation diagram}. This workflow specifically focuses on the in-plane bending motion of a single continuum segment and utilizes limited position and/or orientation measurements from sensors located at strategic points. These sensors include off-the-shelf devices such as Electro-Magnetic (EM) sensors, Inertia Measurement Units (IMU), and cameras, etc., making our approach widely accessible for researchers. The integration of previous shape states and a process model, based on PCK, with measurements from these sensors, feeds into a real-time recursive shape estimation method. This method provides the optimal estimate of the shape state and all corresponding poses along the arc length. Additionally, a comprehensive observability analysis for different sensor configurations is conducted using the noise-weighted observability matrix (detailed in section \ref{ch: observerbility}). An extensive simulation study and experimental validations are discussed in section \ref{ch: simulation} and section \ref{ch: experiment}, respectively. \par

In this work, we focus on in-plane bending to highlight the fundamental concept of polynomial curvature and the derivation of our novel stochastic observer. We note that in-plane bending does not strictly limit the robot to two-dimensional motion: many continuum robot designs can sweep their bending plane $\delta$ in 3D. Thus, while $\delta$ is omitted from our observer state for clarity, it can be added if full 3D reorientation of the bending plane is required in practice. In fact, one can easily place $\delta$ in a separate kinematic module and only add the necessary similarity transformations at the final stage to support more complex 3D motion planning. 

\section{Model Formulation for Observer Design}
\label{ch: model formulation}
This section delineates the development of the shape state, state transition model, and measurement model, all of which are based on PCK. The nomenclature used in these kinematic models is summarized in Table \ref {tab:kinematics nomenclature}. \begin{table}[!h]
  \centering
  \footnotesize
  \caption{Kinematics Nomenclature}
  \begin{tabular}{p{0.14\columnwidth} p{0.80\columnwidth}}
    \thickhline{1.1pt}
    \textbf{Symbol} & \textbf{Description} \\
    \thickhline{0.4pt}
    $s \in [0,1]$ 
      & Normalized arc-length coordinate of one continuum segment \\[4pt]

    $L$ [mm] 
      & Central backbone length of continuum segment \\[4pt]

    $l$ [mm] 
      & Arc-length coordinate at location $s$ \\[4pt]

    $\kappa(s, t)$ 
      & Curvature at location $s$ and time $t$ \\[4pt]

    $m_i(t)$ 
      & Modal coefficient of order $i$ at time $t$ \\[4pt]

    $\theta(s, t)$ 
      & Bending angle at location $s$ and time $t$ \\[4pt]

    $\theta_e$  
      & Bending angle at the segment tip, $\theta_e = \theta(1,t)$ \\[4pt]

    $\delta(t)$ & Bending plane angle at time $t$ \\[4pt]
    
    $\boldsymbol{\psi}$ & Robot state defined by configuration space variables \\[4pt]
    
    $\mb{m}$ & Shape state defined by the modal coefficients vector \\[4pt]
    
    $\mb{J}_{\boldsymbol{\psi}\mb{m}}$ 
      & \textit{Configuration-to-shape} space Jacobian \\[4pt]

    $\mb{J}_{\mb{hm}}$ 
      & \textit{Task-to-shape} space Jacobian \\[4pt]

    Frame\{F\}
      & Right-handed frame with unit vectors 
        $\uvec{x}_f,\uvec{y}_f,\uvec{z}_f$ \\[4pt]

    Frame\{B\}
      & Base frame with origin $\mb{b}$ at the center of the base disk, 
        $\uvec{x}_b$ passing through the first secondary backbone, 
        and $\uvec{z}_b$ perpendicular to the base disk \\[4pt]

    Frame\{1\}
      & Frame obtained by a rotation of $(-\delta)$ about $\hat{\mb{z}}_b$ from Frame \{B\} to characterize the bending plane \\[4pt]

    Frames \{Es\}\& \{Gs\}
      & Frame \{Es\} and Frame \{Gs\} are attached to the axial center of the central backbone at length $s$. The $x$-axis of \{Es\} is the intersection of the bending plane and the orthogonal plane of the central backbone. Frame \{Gs\} is obtained by a rotation $\delta(s,t)$ about $\hat{\mb{z}}_{e_s}$ from \{Es\}. \\[4pt]

    Frames \{E\}\& \{G\}
      & Frame \{E\} and Frame \{G\} are attached to the top surface of the end disk with the unit vectors $\uvec{x}_e,\uvec{y}_e,\uvec{z}_e$ and $\uvec{x}_g,\uvec{y}_g,\uvec{z}_g$. These correspond to \{Es\} and \{Gs\} 
        when $s=1$ \\
    \thickhline{1.1pt}
  \end{tabular}
  \label{tab:kinematics nomenclature}
\end{table}

\subsection{Polynomial Curvature Kinematics} \label{ch:pc_kiematics}
Before we delve into the kinematics model, it's crucial to establish some assumptions to contextualize this problem. The following assumptions are posited:

\begin{enumerate}[(1)]
    \item \textit{Planar Bending Assumption}: The robot's bending motion is strictly planar, negating any out-of-plane deformations.
    \item \textit{Inextensibility of Backbone}: The robot's backbone length remains invariant regardless of its bending configuration.
    \item  \textit{Constant Flexural Rigidity of Backbone}: The robot's shape has a generally smooth curvature profile.
    \item \textit{Disturbance-Agnostic}: External/internal forces (e.g., friction, loads) are treated as probabilistic uncertainties, avoiding explicit modeling.
\end{enumerate}\par
A precise representation and model of the shape for continuum robots(\cite{della2019control},\cite{cheng2022orientation}) is rooted in PCK.
Leveraging this model, all the poses spanning the arclength can be ascertained by integrating a curvature function. The general curvature function is described as an infinite expansion of monomials: \par
\begin{equation}
\label{eqn: infinite poly}
\kappa(s, t)=\sum_{i=0}^{\infty} m_i(t) s^i, \quad s\triangleq \frac{l}{L}, \quad s\in[0,1]
\end{equation}
where  $s$ is a normalized arclength coordinate, $t$ is the time, $m_i$ represents the modal coefficient of $i^\text{th}$-order, $l$ denotes the arclength, and $L$ is the constant central-backbone length.\par
\subsubsection{Finite Dimension Approximation}
\par A finite-dimensional approximation of equation \ref{eqn: infinite poly} with highest order $j$ can be expressed as:
\begin{equation}
\kappa(s, t) \simeq \sum_{i=0}^j m_i(t) s^i
\end{equation}
\par The accuracy of representing a continuum robot's shape is closely related to the order of the curvature function employed. A higher-order curvature function can capture the robot's shape with greater precision. However, this comes at the cost of increased complexity due to the inclusion of additional high-order terms.
\subsubsection{Forward kinematics of configuration-to-task space}
The bending angle $\theta(s,t)$ at the normalized arclength $s$ can be obtained by integrating the above curvature equation.\par
\begin{equation}
\label{eqn: bending angle exp}
\theta(s, t)=\int_0^s \kappa(\tau, t) \mathrm{d}\tau=\sum_{i=0}^{j} m_i(t) \frac{s^{i+1}}{i+1} 
\end{equation}
\par 

\begin{figure}[!b]
	\centering
	\includegraphics[width=1\columnwidth]{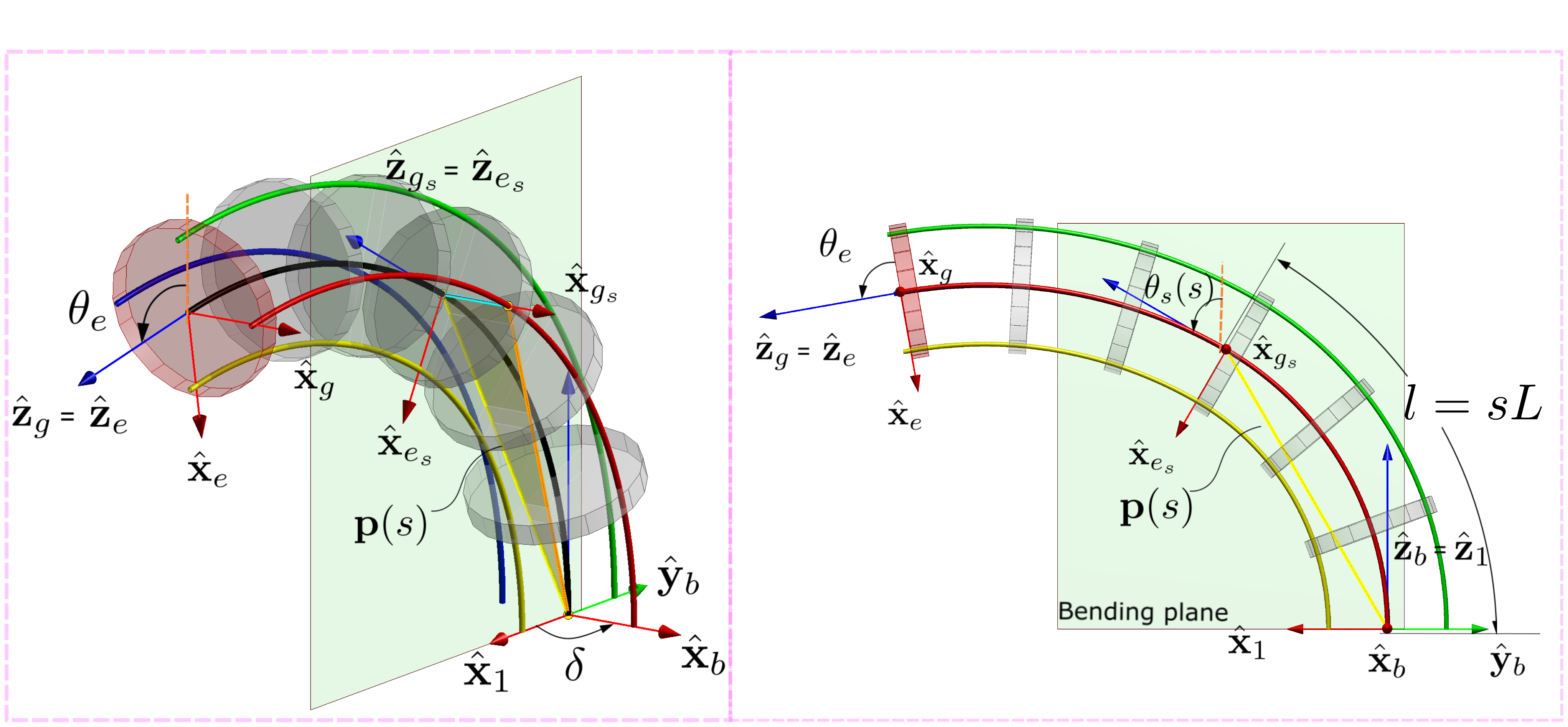}
	\caption{Schematic of  kinematics}
	\label{fig:Direct kinematics}
\end{figure}

In this paper, we follow the frame notation of \cite{wang2019geometric} to derive 
the kinematics (illustrated in Fig.~\ref{fig:Direct kinematics}), and we list the 
definitions of all frames in Table~\ref{tab:kinematics nomenclature}. Because we 
assume no torsional twist along the backbone, the robot’s shape can be fully 
described in a single plane (see the right side of 
Fig.~\ref{fig:Direct kinematics}). Accordingly, only one bending angle 
$\delta(t)$ is required to capture the in-plane motion, with Frame~$\{1\}$ having 
its axes $\uvec{x}_1$ and $\uvec{z}_1$ aligned with the bending plane. Unless 
stated otherwise, all vectors and parameters are expressed in Frame~$\{1\}$.

Let $\mb{p}(s)$ be the position at normalized arclength $s$ in task space.
The tangent unit vector is derived as:
\begin{equation}
\label{eqn: tangent vector}
\mb{e}_t\triangleq\frac{\rmd \mb{p}}{\rmd l} = \frac{\rmd \mb{p}}{\rmd(sL)} = \left[sin\left(\theta(s,t)\right), 0, cos\left(\theta(s,t)\right)\right]\T
\end{equation}
\par Then, the position vector $\mb{p}(s)$ is expressed by the integral of $\frac{\rmd \mb{p}}{\rmd s}$ along the normalized arclength $s$:
\begin{equation}
\label{eqn: position_in_frame1}
\begin{aligned}
   \mb{p}(s) & \triangleq \big[p_x(s), p_y(s), p_z(s)\big]\T \\[6pt]
    & =\int_0^s L \left[sin\left(\theta(\tau,t)\right), 0, cos\left(\theta(\tau,t)\right)\right]\T \rmd \tau 
\end{aligned} 
\end{equation}
\par Combining with the bending angle $\theta(s,t)$, the bending direction angle $\delta(t)$ and the position vector $\mb{p}(s)$, the homogeneous transformation at location $s$ can be obtained:

\begin{equation}
\label{eqn:homo transform}
{ }^{b} \mathbf{T}_{g_s}(s,t)={ }^{b} \mathbf{T}_{1}{ }^{1} \mathbf{T}_{e_s}{ }^{e_s} \mathbf{T}_{g_s}
\end{equation}
where
\begin{align}
&{ }^{b} \mb{T}_{1} =\left[\begin{array}{cc}
\operatorname{RotZ}(-\delta(t)) & \mb{0} \\
\mb{0} & 1
\end{array}\right] \\
&{ }^{1} \mb{T}_{e_s} =\left[\begin{array}{cc}
\operatorname{RotY}(\theta(s,t)) &  \mb{p}(s) \\
\mb{0} & 1
\end{array}\right] \\
&{ }^{e_s} \mb{T}_{g_s} =\left[\begin{array}{cc}
\operatorname{RotZ}(\delta(t)) & \mb{0} \\
\mb{0} & 1
\end{array}\right]
\end{align}
\par The homogeneous transformation in \eqref{eqn:homo transform} follows a sequence 
of rotations and translations between the base frame $\{B\}$, 
Frame~$\{1\}$, and the segment frames $\{E_s\}$ and $\{G_s\}$. For instance, 
Frame~$\{E_s\}$ is obtained from $\{1\}$ by a rotation of $\theta(s,t)$ about 
$\uvec{y}_1$ and a translation by $\mathbf{p}(s)$, while Frame~$\{G_s\}$ is 
obtained by rotating $\{E_s\}$ around $\uvec{z}_{e_s}$ by $\delta(t)$.
\begin{figure}[!t]
	\centering
	\includegraphics[width=1\columnwidth]{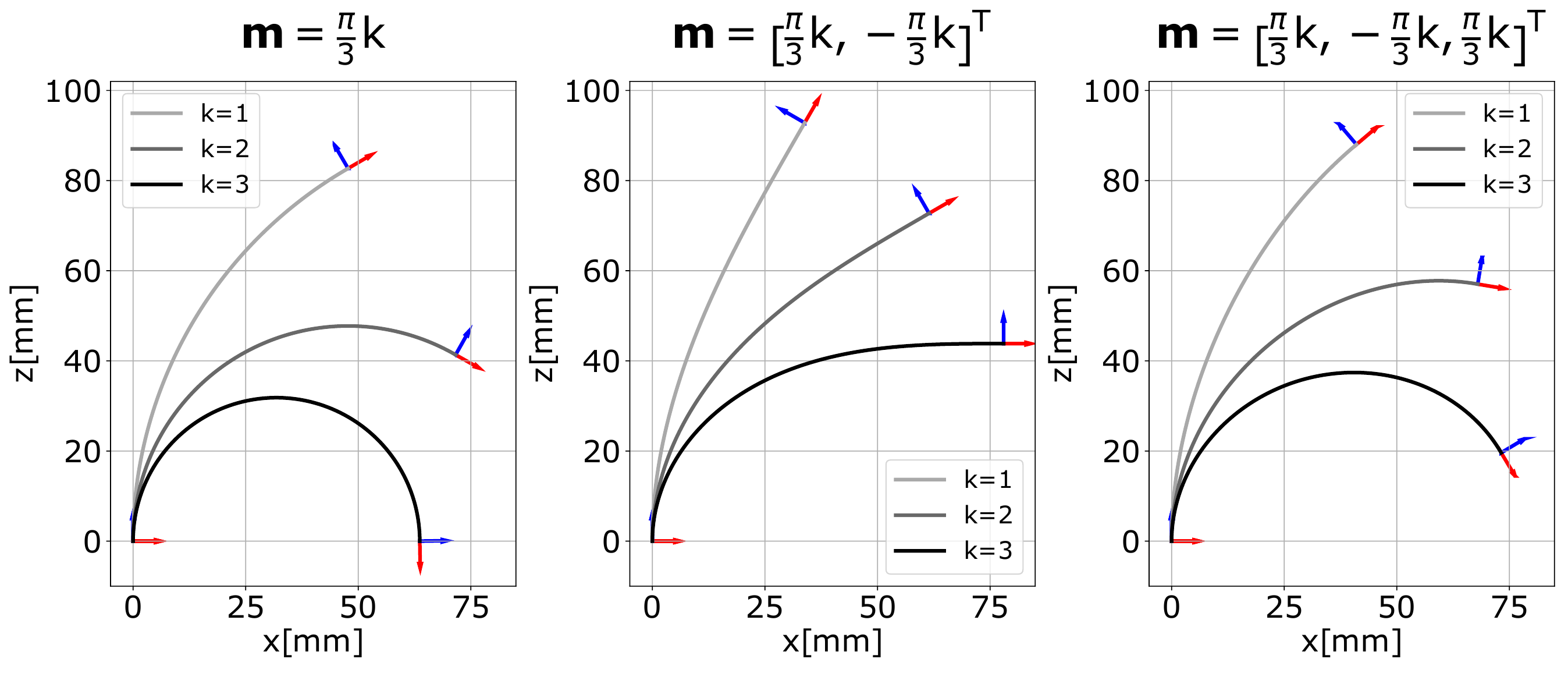}
	\caption{Bending shapes of a continuum robot's central backbone (length $L = 100$ $mm$) generated using $0^{th}$, $1^{st}$ and $2^{nd}$ order polynomial shape state $\mb{m}$. (\textit{Left to Right}: Higher-order polynomials (e.g., $2^{nd}$ order) enable non-uniform curvature profiles with varying curvature rates. \textit{Top to Bottom}: Increasing $k=1,2,3$ amplifies curvature magnitude, leading to sharper bends for each polynomial order.)}
	\label{fig: polynomial curvature shape representation}
\end{figure}

\subsection{Shape State Representation}
In conventional continuum robot modeling, the state is typically represented by configuration space variables such as the tip bending angle \(\theta_e\) and the bending plane angle \(\delta\), based on a constant curvature model:
\begin{equation}
\boldsymbol{\psi} \triangleq\left[\theta_e, \delta\right]\T
\end{equation}\par
However, this model assumes uniform curvature along the robot's length and simplifies the representation, which often does not accurately reflect the complex shape dynamics in environments with disturbances.
\removecomment{In many practical deployments (e.g., endoluminal procedures or constrained workspaces), deformation is predominantly in-plane, and in-plane shape sensing is therefore critical for navigation and control.}{}\editcomment{In many constrained or quasi-planar workspaces, such as bench-top manipulation tasks, continuum arms operating within planar cavities, or tendon-driven instruments designed with torsionally stiff backbones, deformation primarily occurs in a single plane. In these contexts, accurate in-plane shape sensing remains essential for navigation and control, while small out-of-plane deflections can be treated as process disturbances within the stochastic framework. Accordingly, this study focuses on planar bending to establish and validate the core observer formulation before extending it to spatial cases in future work.}{R1-1} Contemporary systems and surveys report continued use of CC/PCC pipelines in applied settings \cite{gunderman2023non,qiu2025actuator,rogatinsky2023multifunctional,russo2023continuum}, while hardware strategies that raise torsional stiffness (thereby suppressing twist in normal use) further support effectively planar operation \cite{santoso2021origami,sun2023enhancing}.
Moreover, in multi-segment tendon-driven continuum robots the bending plane angle \(\delta\) can be directly computed from cable displacements using established kinematic models \cite{camarillo2008mechanics} and thus does not need to be estimated as an additional state variable. In this work, since we focus exclusively on in-plane bending motions, we omit the bending plane angle \(\delta\) from our analysis, thereby simplifying the model to better suit our specific research needs. 
We intentionally exclude the plane angle $\delta$ from our shape state to maintain a clear focus on how polynomial curvature coefficients map directly to sensor measurements. Incorporating $\delta$ requires augmenting every matrix and Jacobian with rotation operators, significantly increasing the notational overhead. This extra layer, though standard, can obscure the core novelty of using non-constant polynomial curvature within a stochastic observer. By omitting $\delta$, we simplify our presentation of the observer’s theoretical development while still enabling 3D motion in real applications by commanding $\delta$ externally, if desired.\par
To enhance accuracy and adaptability in representing the robot's shape, we adopt a modal coefficients representation. By defining the shape state vector \(\mb{m}\) as:
\begin{equation}
\label{eqn: shape_state}
\mb{m} \triangleq \left[ m_0, \dots, m_j\right]\T, \quad
\mb{m} \in \mathbb{R}^{(j+1) \times 1}
\end{equation}
\par
We introduce a scalable method where increasing the order \(j\) of modal coefficients allows for a more precise depiction of the robot's in-plane shape. This approach enhances the fidelity of the shape representation and provides comprehensive shape information, thereby surpassing the accuracy achievable with just the tip bending angle. Some sample shapes represented by \(0^{th}\), \(1^{st}\) and \(2^{nd}\) order polynomials are demonstrated in Fig. \ref{fig: polynomial curvature shape representation}.\par

\subsection{State Transition Model}\label{ch:state_transition}
The scope of our work is confined to evaluating our observer-based shape estimation framework. Consequently, we do not incorporate advanced control strategies but instead opt for a straightforward open-loop control. For simplicity—and without loss of generality—we designate the control input $\mb{u}$ as the change in the bending angle at the tip, denoted by $\delta \theta_e$. This choice is supported by previous studies, such as Camarillo et al.\cite{camarillo2008mechanics}, which demonstrate that in tendon-driven continuum manipulators, the tip bending angle is closely linked to the actuator commands. Similarly, Peng et al.\cite{peng2025dexterous}  successfully used the bending angle as a control parameter in an aerial continuum manipulator, illustrating that this input is not only practical but also adaptable across different application domains. Moreover, our IMM-EKF framework is inherently flexible: once a reliable mapping between the chosen control input and the shape state (modal coefficients) is established, the estimation process can accommodate alternative control inputs if desired.\par

To establish the state transition model, it is essential to delineate the relationship between the change in shape state and the bending angle. By setting $s = 1$, the relationship is expressed through the following mapping derived from equation (\ref{eqn: bending angle exp}):
% \begin{equation}
%     \dot{\theta_e} = \dot{\mb{m}}\T\mb{B}
% \end{equation}
\begin{align}
    \delta \theta_e &= \mb{J}_{\boldsymbol{\psi} \mb{m}} \delta \mb{m} \\
    \mb{J}_{\boldsymbol{\psi} \mb{m}} &= 
    \frac{\partial{\theta(1,t)}}{\partial{\mb{m}}}=\left[ 1, \frac{1}{2}, ..., \frac{1}{j+1}\right]
\end{align}
where $\mb{J}_{\boldsymbol{\psi} \mb{m}} $ is \textit{configuration-to-shape} space Jacobian, facilitating the mapping from modal coefficients to bending angle changes. \par
Subsequently, the infinitesimal change in the shape state is given by:
\begin{equation}
    \delta\mb{m} = \mb{J}_{\boldsymbol{\psi} \mb{m}}^{\dagger} \delta \theta_e
\end{equation}
Here, $\mb{J}_{\boldsymbol{\psi} \mb{m}}^{\dagger}$ represents the Moore-Penrose pseudoinverse of $\mb{J}_{\boldsymbol{\psi}\mb{m}}$, computed as: $\mb{J}_{\boldsymbol{\psi} \mb{m}}^{\dagger} = \mb{J}_{\boldsymbol{\psi} \mb{m}}\T\left(\mb{J}_{\boldsymbol{\psi} \mb{m}}\mb{J}_{\boldsymbol{\psi} \mb{m}}\T\right)^{-1}$. \par
The discrete-time model for the state transition, used at time step $k$, is thus formulated as: 
\begin{equation}
\begin{aligned}
    \mb{m}_k &= \mb{f}\left( \mb{m}_{k-1}, \mb{u}_k\right)  = \mb{m}_{k-1} + \delta \mb{m}_k \\
    \delta \mb{m}_k &= \mb{J}_{\boldsymbol{\psi} \mb{m}}^{\dagger} {\delta \theta_e}_k
     = \mb{J}_{\boldsymbol{\psi} \mb{m}}^{\dagger}\mb{u}_k
\end{aligned}
\end{equation}
This equation represents the discrete-time format of the state transition model, as stipulated in equation (\ref{eqn:problem_formulation_process}).
\subsection{Measurement Model}\label{ch:measurement}
This section outlines the measurement model that integrates the in-plane position and bending angle for a pose sensor located at location $s$ and time $t$. The model is formulated as:
\begin{equation}
\begin{aligned}
\mb{y}_p(s,t) & = \left[ p_x(s,t), p_z(s,t), \theta(s,t)\right]\T \\
&= \left[L\int_0^s sin(\theta(\tau, t))\rmd \tau, L\int_0^s cos(\theta(\tau, t))\rmd \tau, 
\theta(s,t)\right]\T 
\end{aligned}
\label{eqn:pose_meas}
\end{equation}
where $( p_x, p_z)$ denotes the $\mb{x}$ and  $\mb{z}$ coordinates of the sensor in Frame $\{\mathrm{1}\}$. \par
Considering the variability in the number and location of sensors, the measurement model is extended to accommodate the entire array of sensors, thus being represented as a function of the shape state $\mb{m}$:
\begin{equation}
\begin{aligned}
  \mb{h}_p\left(\mb{m}\right) &\triangleq  \left[\mb{y}_p(s_0,t)\T, ... , \mb{y}_p(s_n,t)\T\right]\T, \quad 
  \mb{h}_p \in \mathbb{R}^{3(n+1) \times 1}
\end{aligned}
\end{equation}
where $n+1$ represents the number of sensors. \par
To support the shape estimation framework and given that the measurements are in \textit{task} space, we derive the \textit{task-to-shape} space Jacobian, denoted as $\mb{J}_{\mb{hm}}$.
To simplify its final expression, we introduce a function $\mb{G}(s,t)$ as:
\begin{equation}
\begin{aligned}
    \mb{G}(s,t) &\triangleq \frac{\partial{\mb{y}_p(s,t)}}{\partial{\mb{m}}}
     , \quad \mb{G}(s,t) \in \mathbb{R}^{3 \times (j+1)} \\
    &= \left[\begin{array}{ccccc}
    \frac{\partial{p_x(s,t)}}{\partial{m_0}}
    & \hdots
    &\frac{\partial{p_x(s,t)}}{\partial{m_i}}
    & \hdots
    &\frac{\partial{p_x(s,t)}}{\partial{m_j}}\\[6pt]
    \frac{\partial{p_z(s,t)}}{\partial{m_0}}
    & \hdots
    & \frac{\partial{p_z(s,t)}}{\partial{m_i}}
    & \hdots
    & \frac{\partial{p_z(s,t)}}{\partial{m_j}} \\[6pt]
    \frac{\partial{\theta(s, t)}}{\partial{m_0}} 
    & \hdots
    &\frac{\partial{\theta(s, t)}}{\partial{m_i}} 
    & \hdots
    &\frac{\partial{\theta(s, t)}}{\partial{m_j}} 
    \end{array}\right]   
\end{aligned}
\end{equation}
where 
\begin{align}
\frac{\partial{p_x(s,t)}}{\partial{m_i}}  &=L\int_0^scos(\theta(\tau, t))\frac{\tau^{i+1}}{i+1} \rmd \tau \\
% line 6
\frac{\partial{p_z(s,t)}}{\partial{m_i}}  &=L\int_0^s-sin(\theta(\tau, t))\frac{\tau^{i+1}}{i+1} \rmd \tau  \\
\frac{\partial{\theta(s, t)}}{\partial{m_i}} &= \frac{s^{i+1}}{i+1}
\end{align}
Finally, $\mb{J}_{\mb{hm}}$ is defined as:
\begin{equation}
\label{eqn:identification Jacobian}
\begin{aligned}
\mb{J}_{\mb{hm}} &\triangleq \frac{\partial \mb{h}_p\left( \mb{m}\right)}{\partial \mb{m}} , \quad \mb{J}_{\mb{hm}} \in \mathbb{R}^{3(n+1) \times (j+1)} \\
& = \left[
\mb{G}(s_0,t)\T, ... ,
\mb{G}(s_n,t)\T
\right]\T
\end{aligned}
\end{equation}
\par 
Adjustments to the measurement model may be necessary depending on the specific sensor types used (position or orientation sensors) and their configurations. For example, equation \eqref{eqn:pose_meas} becomes $\mb{y}_p(s,t) = \left[ p_x(s,t), p_z(s,t)\right]\T$ for a position sensor and $\mb{y}_p(s,t) = \left[ \theta(s,t)\right]\T$ for an orientation sensor, respectively. Variations in sensor arrangement necessitate corresponding modifications in the measurement model and its Jacobian matrix.
% %
% \input{content/model_formulation_for_observer_design_v2}
\section{IMM-EKF Design}
\label{ch:recursive_est_desgin}
The EKF is a robust, real-time state estimation technique for nonlinear systems, widely acclaimed for its effectiveness. However, the complex dynamic behaviors of continuum robots often render single-model approaches inadequate. The IMM method addresses this limitation by enabling seamless transitions between multiple models, thereby capturing the evolving shape of continuum robots in various situations. \par
In this study, we utilize three polynomial curvature models of different orders: $0^{th}$ order, $1^{st}$ order, and $2^{nd}$ order, positing that the $2^{nd}$ order polynomial provides the highest fidelity. Although the $2^{nd}$ order polynomial offers greater precision, it may overfit and exhibit increased sensitivity to noise in simpler environments where a lower-order model could prove more robust. \par
While the standard IMM-EKF algorithm is commonly used, it has two notable limitations that motivated our refinements:
\begin{itemize}
    \item \emph{Dimension Mismatch:} The polynomial curvature models differ in state-vector size ($0^{the}, 1^{st}$, and $2^{nd}$ order). A naive approach that merely appends zeros to smaller-dimensional states can bias the mixed initial states, since these appended zeros might be interpreted as “known” values rather than unknown parameters. 
    \item \emph{Fixed Transition Probability Matrix (TPM):} Continuum robots often experience abrupt changes in shape due to dynamic external interactions. The standard IMM relies on a fixed TPM, defined \emph{a priori}, which assumes stationary transition probabilities between modes. This static assumption leads to delayed adaptation during sudden transitions (if TPM entries are underconfident) or overfitting to spurious mode changes (if TPM entries are overconfident).
\end{itemize}
In the subsequent sections, we revisit the CIMM method and introduce our refined approach. We address two key challenges in the model interaction stage to enhance overall performance.
The main nomenclature for the observer design is illustrated in Table ~\ref{tab:imm_ekf nomenclature}.
\begin{table}[!t]
  \centering
  \footnotesize
  \caption{IMM-EKF Nomenclature}
  \begin{tabular}{p{0.14\columnwidth} p{0.80\columnwidth}}
    \thickhline{1.1pt}
    \textbf{Symbol} & \textbf{Description} \\
    \thickhline{0.4pt}

    $M^{(i)}$ 
      & System model using the $i^\text{th}$-order polynomial curvature function \\[4pt]
      
    $\check{V}_k$ 
      & Predicted value of variable \(V\) at time \(k\), given measurements up to time \(k-1\) \\[4pt]
      
    $\hat{V}$ 
      & Estimated value of variable \(V\) after the correction stage of filtering \\[4pt]

    $\mb{P}$ 
      & Markov probability transition matrix \\[4pt]

    $p_{ij}$ 
      & Probability of transitioning from \(M^{(i)}\) at time \(k-1\) to \(M^{(j)}\) at time \(k\) \\[4pt]

    $P(A \mid B)$ 
      & Probability of event \(A\), given event \(B\) \\[4pt]

    $\mu^{(j)|(i)}_{k-1}$ 
      & Mixing probability at time \(k-1\) for \(M^{(j)}\), given that \(M^{(i)}\) was correct at time \(k-1\) \\[4pt]

    $\mu^{(i)}_k$ 
      & Probability of \(M^{(i)}\) at time \(k\), given all observations up to time \(k\) \\[4pt]

    $\check{\mu}^{(i)}_k$ 
      & A priori (predicted) model probability at time \(k\) for \(M^{(i)}\), given observations up to time \(k-1\) \\[4pt]

    $\hat{\mb{m}}_{k-1}^{(0i)}$ 
      & Mixed initial state vector of \(M^{(i)}\) \\[4pt]

    $\hat{\mb{C}}_{k-1}^{(0i)}$ 
      & Mixed initial covariance matrix of \(M^{(i)}\) \\[4pt]

    $\Lambda^{(i)}_k$ 
      & Probability of observing \(\mb{y}_k\), given all previous observations and assuming \(M^{(i)}\) is correct \\[4pt]

    $\mathcal{N}\bigl(x;\mu,\Sigma\bigr)$ 
      & Gaussian probability density function with mean \(\mu\) and covariance \(\Sigma\) \\[4pt]

    $\bs{\nu}_\mb{r}$ 
      & Simulated position noise \\[4pt]

    $\bs{\nu}_\mb{\psi}$ 
      & Simulated bending angle noise \\[4pt]

    $\hat{\mb{m}}_0^j$ 
      & Initial estimated state of \(M^{(j)}\) \\[4pt]

    $\hat{\mb{C}}_{0}^j$ 
      & Initial state covariance of \(M^{(j)}\) \\[4pt]

    $\mb{Q}$ 
      & State process covariance \\[4pt]

    $\mb{R}$ 
      & State measurement covariance \\[4pt]

    $\mb{F}_k^{(i)}$ 
      & State transition matrix of the \(i^\text{th}\) EKF at time step \(k\) \\[4pt]

    $\mb{H}_k^{(i)}$ 
      & Observation matrix of the \(i^\text{th}\) EKF at time step \(k\) \\[4pt]

    $\mb{S}_k^{(i)}$ 
      & Residual covariance of the \(i^\text{th}\) EKF at time step \(k\) \\[4pt]

    $\mb{K}_k^{(i)}$ 
      & Near-optimal Kalman gain of the \(i^\text{th}\) EKF at time step \(k\) \\

    \thickhline{1.1pt}
  \end{tabular}
  \label{tab:imm_ekf nomenclature}
\end{table}

\subsection{IMM Formulation}

\begin{figure}[!h]
  \centering
  \includegraphics[width=1\columnwidth]{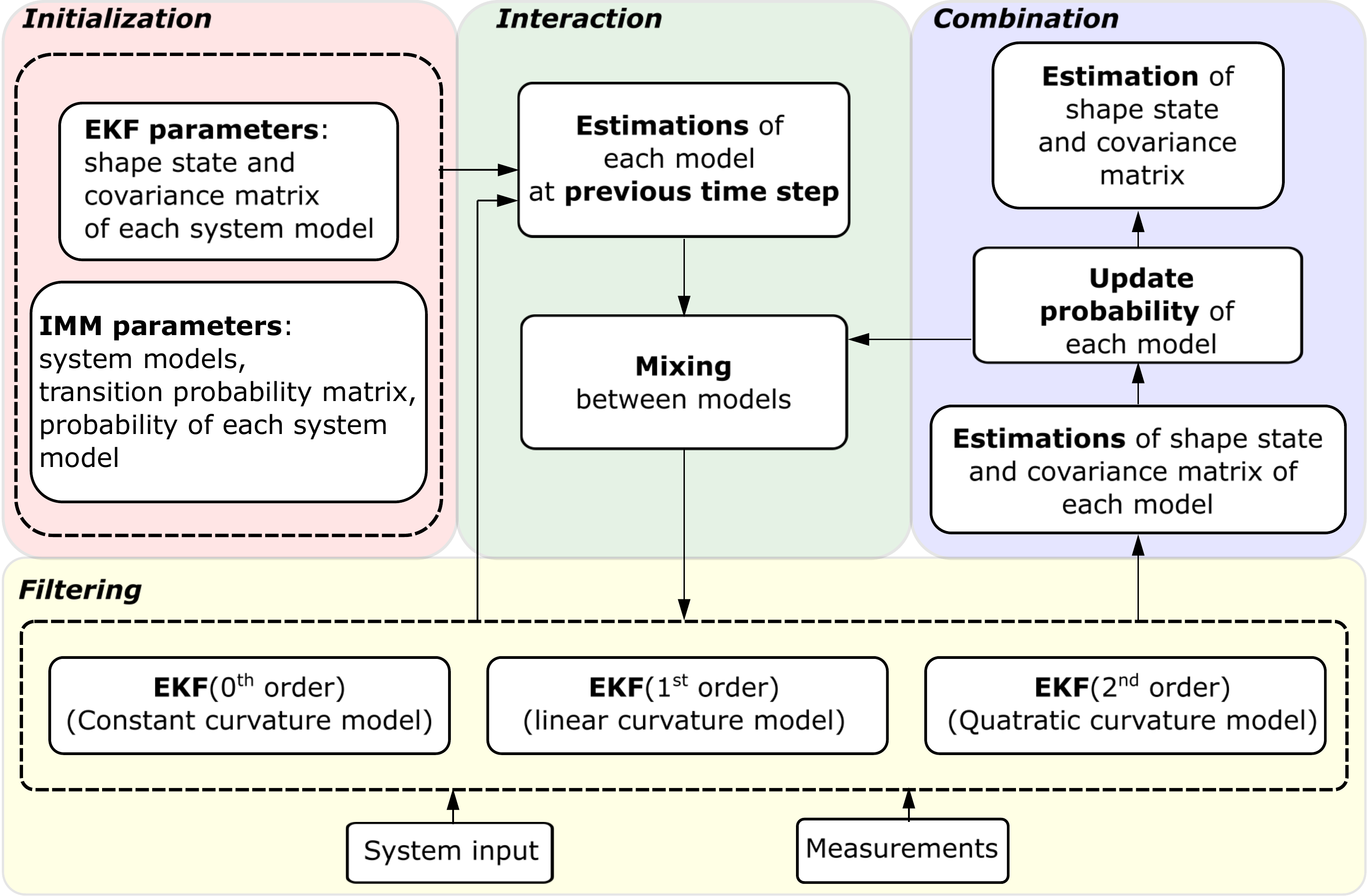}
  \caption{Schematic of IMM-EKF algorithm}
  \label{fig:IMM-EKF diagram}
\end{figure}

Figure~\ref{fig:IMM-EKF diagram} illustrates the three-stage IMM-EKF cycle: \emph{interaction}, \emph{filtering}, and \emph{combination}. During interaction, the state estimates from the previous time step are merged to provide initial conditions for each model’s filter. In the filtering stage, all models simultaneously perform prediction and update. Finally, in the combination stage, the updated state estimates from each model are fused into a single state estimate, weighted by the updated model probabilities. Detailed mathematical derivations for each stage are provided in the following subsections.

\subsubsection{Multiple Model Set}

A system model is denoted by $M^{(i)} \in \mathbb{M} = \{M^{(0)}, M^{(1)}, \ldots, M^{(r)}\}$, where the superscript $i$ refers to a polynomial curvature function of order $i$. The discrete-time process and measurement models for 
$M^{(i)}$ are:
\begin{equation}
\begin{aligned}
\mb{m}_{k}^{(i)} & = \mb{f}\bigl(\mb{m}_{k-1}^{(i)}, \mb{u}_{k-1}\bigr) 
+ \mb{v}_{k-1}^{(i)} \\
\mb{y}_k & = \mb{h}_{p}\bigl(\mb{m}_{k}^{(i)}\bigr) + \mb{n}_{k}^{(i)}
\end{aligned}
\label{eqn: state measurement dynamics}
\end{equation}
where $\mb{f}$ is the state transition model (see 
Section~\ref{ch:state_transition}), and $\mb{h}_p$ is the measurement model 
(see Section~\ref{ch:measurement}). Here, $\mb{y}_k$ is the measurement at time step $k$, and $\mb{v}_{k-1}^{(i)}$ and $\mb{n}_{k}^{(i)}$ are independent Gaussian noise terms with means $\bar{\mb{v}}$, $\bar{\mb{n}}$ and covariance matrices $\mb{Q}$, $\mb{R}$, respectively. The model set $\mathbb{M}$ is time-invariant. In this work, we use three models (i.e., $r = 2$).

\subsubsection{Probability Transition Matrix}
The Markov transition probability matrix (TPM) $\mb{P}$ describes the likelihood 
of switching from one model to another between consecutive time steps. Each 
entry $p_{ij}$ in $\mb{P}$ is the \emph{a priori} probability of transitioning from model $i$ at time $k-1$ to model $j$ at time $k$. Specifically,
\begin{align}
\mb{P} 
  = \begin{bmatrix}
    p_{00} & \dots & p_{0r} \\
    \vdots & p_{ij} & \vdots \\
    p_{r0} & \dots & p_{rr}
  \end{bmatrix}, 
\quad 
\sum_{j=0}^r p_{ij} = 1,\quad 
0 \leq p_{ij} \leq 1
\end{align}

\subsubsection{State Interaction / Mixing}
At the start of each new time step, \emph{mixing probabilities} set the initial conditions for each model by accounting for possible transitions from all other models. The mixing probability of transitioning from model $j$ to $i$ is
\begin{equation}
\mu_{k-1}^{(j)\mid(i)} = 
\frac{p_{ji}\,\mu_{k-1}^{(j)}}{\check{\mu}_{k}^{(i)}}, 
\quad 
\check{\mu}_{k}^{(i)} 
= \sum_{j=0}^r p_{ji}\,\mu_{k-1}^{(j)}
\end{equation}
(Derivation details are given in Supplementary Appendix B.)

For each model $i$, the mixed initial state and covariance combine the previous estimates and their uncertainties, weighted by the mixing probabilities:
\begin{align}
&\hat{\mb{m}}_{k-1}^{(0i)} =\sum_{j=0}^r \mu_{k-1}^{(j) \mid (i)}  \hat{\mb{m}}_{k-1}^{(j)} \\
& \begin{aligned}
&\hat{\mb{C}}_{k-1}^{(0i)} = \sum_{j=0}^r \mu_{k-1}^{(j) \mid (i)} \Bigl(\hat{\mb{C}}_{k-1}^{(j)} \\
& + \left(\hat{\mb{m}}_{k-1}^{(j)}-\hat{\mb{m}}_{k-1}^{(0i)}\right)\left(\hat{\mb{m}}_{k-1}^{(j)}-\hat{\mb{m}}_{k-1}^{(0i)}\right)\T\Bigl)
\end{aligned}
\label{eqn: cal_mixed_initial_state}
\end{align}

\subsubsection{Mode-Matched Filtering}

Each candidate model then performs prediction and update independently, starting with the mixed initial conditions. This parallel filtering step ensures the IMM maintains multiple hypotheses about the system state and can quickly adapt to new measurements.

\textbf{Prediction} for model $i$:
\begin{align}
\check{\mb{m}}_{k}^{(i)} 
  &= \mb{f}\bigl(\hat{\mb{m}}_{k-1}^{(0i)}, \mb{u}_k\bigr) \\
\check{\mb{C}}_{k}^{(i)} 
  &= \mb{F}_k^{(i)}\,\hat{\mb{C}}_{k-1}^{(0i)}
     \bigl(\mb{F}_k^{(i)}\bigr)\T 
     + \mb{Q}^{(i)}
\end{align}
where $\mb{F}_k^{(i)}$ is the Jacobian of $\mb{f}$ with respect to $\mb{m}$.

\textbf{Correction} for model $i$:
\begin{align}
\mb{S}_k^{(i)} 
  &= \mb{H}_k^{(i)}\,\check{\mb{C}}_{k}^{(i)}
     \bigl(\mb{H}_k^{(i)}\bigr)\T 
     + \mb{R}^{(i)}\\
\mb{K}_k^{(i)} 
  &= \check{\mb{C}}_{k}^{(i)}
     \bigl(\mb{H}_k^{(i)}\bigr)\T
     \bigl(\mb{S}_k^{(i)}\bigr)^{-1} \\
\hat{\mb{m}}_{k}^{(i)} 
  &= \check{\mb{m}}_{k}^{(i)} 
     + \mb{K}_k^{(i)}
       \Bigl(\mb{y}_k 
             - \mb{h}_p\bigl(\check{\mb{m}}_{k}^{(i)}\bigr)\Bigr) \\
\hat{\mb{C}}_{k}^{(i)} 
  &= \bigl(\mb{I} - \mb{K}_k^{(i)}\,\mb{H}_k^{(i)}\bigr)\
     \check{\mb{C}}_{k}^{(i)}
\end{align}

\subsubsection{Model Likelihood and Probability Update}

After filtering, the likelihood of each model $i$ is updated based on the latest 
measurement. Using the multivariate Gaussian distribution, the likelihood is
\begin{equation}
\Lambda_k^{(i)}
 = \mathcal{N}\Bigl(\mb{y}_k 
                    - \mb{h}_p\bigl(\check{\mb{m}}_{k}^{(i)}\bigr);\, 
                    \mb{0},\,\mb{S}_k^{(i)}\Bigr)
\end{equation}
The model probability $\mu_k^{(i)}$ is then computed via Bayes’ rule:
\begin{equation}
\mu_k^{(i)}
  = \frac{\check{\mu}_{k}^{(i)}\,\Lambda_k^{(i)}}
         {\sum_{j=0}^r \check{\mu}_{k}^{(j)}\,\Lambda_k^{(j)}}
\end{equation}
(Derivation details are given in Supplementary Appendix C.)

\subsubsection{Model Combination}

Finally, the updated estimates from all models are fused to produce the overall 
estimate of the system state:
\begin{equation}
\hat{\mb{m}}_{k}
  = \sum_{i=0}^r \mu_k^{(i)}\,\hat{\mb{m}}_{k}^{(i)}
\end{equation}
The combined covariance measures the overall uncertainty:
\begin{equation}
\hat{\mb{C}}_{k}
  = \sum_{i=0}^r \mu_k^{(i)}
    \Bigl[
      \hat{\mb{C}}_{k}^{(i)}  
      + \bigl(\hat{\mb{m}}_{k} 
              - \hat{\mb{m}}_{k}^{(i)}\bigr)\,
        \bigl(\hat{\mb{m}}_{k} 
              - \hat{\mb{m}}_{k}^{(i)}\bigr)\T
    \Bigr]
\end{equation}

\subsection{Refining IMM}
The CIMM approach
is widely used for multi-model state estimation, yet it exhibits
two notable limitations. First, it struggles with the mixing of state vectors with unequal dimensions. In such cases, the mixing stage may introduce biases when smaller-dimensional state vectors are combined with larger ones, potentially leading to inaccuracies in state estimation and model predictions. In multi-model settings for continuum robots, state dimensions differ because each curvature model (e.g., $0^{th}, 1^{st}$, and $2^{nd}$ order polynomials) uses a different number of parameters. Simply forcing smaller vectors to match larger ones can introduce artificial zeros, which may be misinterpreted by the filter as certain knowledge that these polynomial coefficients are exactly zero. This can skew the mixed state estimates, especially if those coefficients in the higher-order model are significant. Second, the conventional IMM uses a fixed TPM predefined based on prior assumptions about system dynamics. This static approach fails to accommodate environmental or internal disturbances that could dynamically alter system behavior, thus limiting IMM's effectiveness in unpredictable scenarios. To handle dynamic shape variations in continuum robots, the TPM must adapt to real-time measurements. If a lower-order model becomes suddenly inadequate (e.g., when the robot bends under more complex external loading conditions), a higher-order model should gain probability quickly. Conversely, if the environment simplifies, the probability of the simpler model should increase. Therefore, we implement an online correction mechanism for $p_{ij}$ to reduce lag errors and better reflect sudden shape changes.\par
To address these issues, we refine the standard IMM-EKF framework by:\par
\subsubsection{Mitigating Bias in State Dimensionality}
In the mixing stage of the IMM, the mixed initial state for each model is derived based on Equation \eqref{eqn: cal_mixed_initial_state}, which requires uniform dimensionality across all state vectors and their corresponding covariance matrices. The conventional IMM approach often addresses dimensional discrepancies by setting augmented components in state vectors with smaller dimensions to zero. This leads to potential biases in the mixed initial states of models with larger dimensions.
Inspired by \cite{yuan2012multiple}, which provides an unbiased mixing solution for two models, we propose an unbiased mixing strategy tailored for multiple models with varying dimensions, illustrated using the case of $\mathbb{M} = \left\{M^{\mathit{(0)}}, M^{\mathit{(1)}}, M^{(2)}\right\}$. \par
Consider the state vector and covariance matrix of $M^{\mathit{(2)}}$ as follows:
\begin{equation}
\hat{\mb{m}}^{\mathit{(2)}}=\left[\begin{array}{c}
\hat{m}_0^{\mathit{(2)}} \\ [3pt]
\hat{m}_1^{\mathit{(2)}} \\ [3pt]
\hat{m}_2^{\mathit{(2)}}
\end{array}\right],  \hat{\mb{C}}^{\mathit{(2)}}=\left[\begin{array}{ccc}
C_{0 }^{\mathit{(2)}} & C_{0  1}^{\mathit{(2)}} & C_{0  2}^{\mathit{(2)}} \\ [3pt]
C_{1 0}^{\mathit{(2)}} & C_{1 }^{\mathit{(2)}} &C_{1 2}^{\mathit{(2)}} \\ [3pt]
C_{2 0}^{\mathit{(2)}} & C_{2 1}^{\mathit{(2)}} &C_{2 }^{\mathit{(2)}} 
\end{array}\right]
\end{equation}
The corresponding state vectors and covariance matrices for $M^{(1)}$ and $M^{(0)}$ are defined similarly but with reduced dimensions.

To ensure dimensional consistency and unbiased mixing, we integrate unique components from the largest state vector into those of smaller dimensions. For example, the component $\hat{m}_2^{(2)}$ specific to $M^{\mathit{(2)}}$ is added to the state vectors of $M^{\mathit{(0)}}$ and $M^{\mathit{(1)}}$. Additionally, where the component $\hat{m}_1$ exists across multiple models (as in $M^{\mathit{(1)}}$ and $M^{\mathit{(2)}}$ ), it is averaged to form the augmented component for $M^{\mathit{(0)}}$. \par
The augmented state vectors and their covariance matrices of $M^{\mathit{(1)}}$  and $M^{\mathit{(0)}}$ are then constructed as follows:

\begin{equation}
\hat{\mb{m}}^{\mathit{(1)}}=\left[\begin{array}{c}
\hat{m}_0^{\mathit{(1)}} \\
\hat{m}_1^{\mathit{(1)}} \\
\hat{m}_2^{\mathit{(2)}}
\end{array}\right] ,  \hat{\mb{C}}^{\mathit{(1)}}=\left[\begin{array}{ccc}
C_0^{\mathit{(1)}} & C_{01}^{\mathit{(1)}} & 0 \\
C_{10}^{\mathit{(1)}} & C_1^{\mathit{(1)}} & 0 \\
0 & 0 & C_2^{\mathit{(2)}}
\end{array}\right] 
\label{eqn: augmented_states_covariance_M1}
\end{equation}

\begin{equation}
\begin{aligned}
\hat{\mb{m}}^{\mathit{(0)}} &= \left[
\hat{m}_0^{\mathit{(0)}},
\frac{\hat{m}_1^{\mathit{(1)}}+\hat{m}_1^{\mathit{(2)}}}{2},
\hat{m}_2^{\mathit{(2)}}
\right]\T, \\
\hat{\mb{C}}^{\mathit{(0)}} &= \textbf{diag}\left(C_0^{\mathit{(0)}}, \frac{C_1^{\mathit{(1)}}+C_1^{\mathit{(2)}}}{4}, C_2^{\mathit{(2)}}\right) 
\end{aligned}
\label{eqn: augmented_states_covariance_M0}
\end{equation}

This approach ensures that each model's state vector in the mixing stage is dimensionally consistent with the most complex model, thereby reducing bias in the IMM's estimations. Note that the augmented components are considered only during the mixing stage. They are excluded from subsequent EKF iterations to maintain focus on the relevant dimensions of each model.

\subsubsection{Designing Adaptive Transition Strategies}
The conventional IMM method often struggles to meet performance expectations in dynamic environments due to its reliance on a fixed TPM predicated on prior knowledge. This limitation has spurred researchers to develop methods that enable the TPM to adapt more effectively, not just based on historical data but also incorporating real-time information. We build upon these efforts in our work, especially drawing on the correction functions proposed in \cite{xie2019adaptive} and \cite{lee2023improved}. We leverage these concepts to refine and implement a TPM adjustment mechanism that responds dynamically to current system states, enhancing the IMM's responsiveness and accuracy in diverse operation scenarios. \par
The first correction function is proposed based on the model probability variation trend. it can be expressed as
\begin{align}
f_k^{(j)} & =\frac{1}{1-\Delta \mu_k^{(j)}}, \quad(j=0,1, \ldots, r) \\
\Delta \mu_k^{(j)} & =\mu_k^{(j)}-\mu_{k-1}^{(j)}
\end{align}
where $\Delta \mu_k^{(j)}$ is the gradient of the probability of model $j$.\par
The element of TPM is corrected by
\begin{equation}
\bar{p}_{i j}(k)=p_{i j}(k-1) f_k^{(j)} , \quad(i=0,1, \ldots, r)
\label{eqn: 1st_corrected_pij}
\end{equation}
The first correction function $f_k^{(j)}$ benefits the system when the true model is unchanged. It cannot improve the responsiveness of the IMM method when a model jump happens. \par
We utilize an activating correction function described in  \cite{lee2023improved} to reduce lag errors and improve the responsiveness of IMM when the model jumps. The activating function is designed as
\begin{align}
g_k^{(i)} & =\frac{1}{1+ \exp^{\Delta^{(i)}}}, \quad(i=0,1, \ldots, r) \\
\Delta^{(i)} & =\mu_k^{(i)}-\sum_{j=1, j \neq i}^r \mu_{k-1}^{(j)}
\end{align}
where $\Delta^{(i)}$ is the cross-difference model probability from model $j$ to model $i$ at time step $k-1$ and $k$. It examines the switching between models. Implementing this correction function to $p_{i j}$ can reduce the delay time in the model transition period. \par
Similar to \eqref{eqn: 1st_corrected_pij}, the corrected term of TPM using $g_k^{(i)}$ is
\begin{equation}
\Tilde{p}_{i j}(k)=p_{i j}(k-1) g_k^{(i)} , \quad(i=0,1, \ldots, r)
\label{eqn: 2nd_corrected_pij}
\end{equation}
Combining \eqref{eqn: 1st_corrected_pij} and \eqref{eqn: 2nd_corrected_pij}, we can obtain an integrated term of TPM as

\begin{equation}
\begin{aligned}
\hat{p}_{i j}(k)&=\bar{p}_{i j}(k) - \Tilde{p}_{i j}(k) , \quad(i=0,1, \ldots, r) \\
&= p_{i j}(k-1) (f_k^{(j)} - g_k^{(i)})
\label{eqn: final_corrected_pij}
\end{aligned}
\end{equation}
To satisfy the requirement of the Markov chain, the sum of each row of TPM is supposed to equal 1. So, the element of corrected TPM is further normalized as
\begin{equation}
p_{i j}(k)=\frac{\hat{p}_{i j}(k)}{\sum_{j=1}^M \hat{p}_{i j}(k) }
\end{equation}
\par Unlike the strategy of using $g_k^{(i)}$ at all time steps in \cite{lee2023improved}, we introduce the minimum likelihood ratio between models as a prior judging condition to know if the current model jumps. The reason is that this correction function may cause instability in cases with large noises and no dominant model. The likelihood ratio is expressed as
\begin{equation}
\lambda_{i j}=\frac{\Lambda_i}{\Lambda_j}, \quad j=[0,1, \ldots, r]
\end{equation}
where $i \neq j$, $\Lambda_i$ and $\Lambda_j$ represent the likelihoods of model $i$ and $j$. \par
For the situation with model $i$ being the matched model, it will have the highest likelihood compared to others. Thus, the minimum likelihood ratio $\min[\lambda_{i j}]$ is larger than 1. When model $i$ becomes unmatched, $\min[\lambda_{i j}]$ will largely decrease and be less than 1. So, $\min[\lambda_{i j}]$ can be used as a status indicator to know whether model jumping happens. For model $i$, We incorporate a model jumping threshold $\mathrm{Th}_i$ with $\min[\lambda_{i j}]$ to judge the system status. \par
Based on the above analysis, the element of corrected TPM can be further updated as
\begin{equation}
\hat{p}_{i j}(k)= \begin{cases}\bar{p}_{i j}(k) - \Tilde{p}_{i j}(k), & \min \left[\lambda_{i j}\right] \leq \mathrm{Th}_{i} \\ \bar{p}_{i j}(k) , & \min \left[\lambda_{i j}\right]>\mathrm{Th}_{i}\end{cases}
\label{eqn: dual_correction}
\end{equation}
\subsection{Proposed Algorithm}
We can simplify the strategies for state augmentation and TPM correction to mitigate the risk of overestimation and instability. \par
By assuming zero covariance on extra components of the augmented state vector,  \eqref{eqn: augmented_states_covariance_M1} and   \eqref{eqn: augmented_states_covariance_M0} become
\begin{equation}
\hat{\mb{m}}^{\mathit{(1)}}=\left[\begin{array}{c}
\hat{m}_0^{\mathit{(1)}} \\ [3pt]
\hat{m}_1^{\mathit{(1)}} \\ [3pt]
\hat{m}_2^{\mathit{(2)}}
\end{array}\right] , \quad \hat{\mb{C}}^{\mathit{(1)}}=\left[\begin{array}{ccc}
C_0^{\mathit{(1)}} & C_{01}^{\mathit{(1)}} & 0 \\ [3pt]
C_{10}^{\mathit{(1)}} & C_{1}^{\mathit{(1)}} & 0 \\ [3pt]
0 & 0 & 0
\end{array}\right] \\
\label{eqn: simplified_augmented_states_covariance_M1}
\end{equation}
\begin{equation}
\begin{aligned}
\hat{\mb{m}}^{\mathit{(0)}} &= \left[
\hat{m}_0^{\mathit{(0)}},
\frac{\hat{m}_1^{\mathit{(1)}}+\hat{m}_1^{\mathit{(2)}}}{2},
\hat{m}_2^{\mathit{(2)}}
\right]\T, \\
\hat{\mb{C}}^{\mathit{(0)}} &= \textbf{diag}\left(C_0^{\mathit{(0)}}, 0, 0\right) 
\end{aligned}
\label{eqn: simplified_augmented_states_covariance_M0}
\end{equation}
Note that while zero covariance in the extra augmented components may seem to imply complete certainty, here it serves as a modeling convenience to “reset” certain states when switching modes. This prevents overestimation and runaway cross-correlations but does not eliminate the usual uncertainty that is reintroduced during the subsequent filter update steps.\par
By ignoring the second correction function $g_k^{(i)}$, \eqref{eqn: dual_correction} can be simplified as
\begin{equation}
\hat{p}_{i j}(k)=\bar{p}_{i j}(k)
\label{eqn: single_correction}
\end{equation}
\par Based on theoretical considerations and preliminary analysis, we initially conceptualize two enhanced IMM-EKFs based on these refining strategies. \par
\begin{itemize}
    \item \textbf{Dual Adaptive IMM-EKF (DAIMM-EKF)}: This approach assumes zero covariance for the augmented state vectors using \eqref{eqn: simplified_augmented_states_covariance_M1} and \eqref{eqn: simplified_augmented_states_covariance_M0}, and employs a dual-correction function \eqref{eqn: dual_correction} for TPM modulation. This combination can quickly adapt to abrupt changes and reduce lag during model jumps, but requires careful tuning and thresholds to avoid instability under noisy conditions.
    \item \textbf{Single Adaptive IMM-EKF (SAIMM-EKF)}: This approach maintains nonzero covariance based on the original state elements for the augmented states using \eqref{eqn: augmented_states_covariance_M1} and \eqref{eqn: augmented_states_covariance_M0}, and uses a single correction function \eqref{eqn: single_correction} for TPM adjustment. This approach is often more conservative in model switching, potentially avoiding instability risks from dual adaptation in noisy scenarios.
\end{itemize} \par
The entire process of these enhanced IMM-EKFs is detailed in Algorithm~\ref{algo:IMM-EKF AL}. 

\begin{algorithm}
\caption{Proposed enhanced IMM-EKF method}
 \label{algo:IMM-EKF AL}
\begin{algorithmic}[1]
 \small
 \Require $\mathcal{D}\{\mb{y}_k,\mb{u}_k\}$, $_{k=1,\hdots,N}$; $\mb{Q}_{\bar{\mb{v}}}$; $\mb{R}_{\bar{\mb{n}}}$; 

 \item[]
 
\item[] \textbf{Initialization:}
$\mb{P}_0$, $\mu_0^{(j)}$,
$\hat{\mb{m}}_0^{(j)}$,$\hat{\mb{C}}_{0}^{(j)}$ $\quad j=[0,1, \ldots, r]$
\For{$k=1 \; \textbf{to} \; N$}
\item[]

\item[] \textbf{1) Model interaction:}
    \For{$i=0 \; \textbf{to} \; r$}
        \State \( \check{\mu}_k^{{(i)}} = \sum_{j=0}^{r} p_{ji} \mu_{k-1}^{{(j)}} \)
        \State \( \mu_{k-1}^{{(j)} \mid {(i)}} = \frac{p_{j i} \mu_{k-1}^{{(j)}}}{\check{\mu}_{k}^{{(i)}}}  \)
        \State \textbf{Update $\hat{\mb{m}}_{k-1}^{j}$ and $\hat{\mb{C}}_{k-1}^{{(j)}}$ using \eqref{eqn: simplified_augmented_states_covariance_M1}, \eqref{eqn: simplified_augmented_states_covariance_M0}  or \eqref{eqn: augmented_states_covariance_M1}, \eqref{eqn: augmented_states_covariance_M0}} 
        \State \( \hat{\mb{m}}_{k-1}^{(0i)} = \sum_{j=0}^{r} \mu_{k-1}^{{(j)} \mid {(i)}} \hat{\mb{m}}_{k-1}^{{(j)}} \)
        \State $\hat{\mb{C}}_{k-1}^{(0i)}=\!
        \begin{aligned}[t]
           & \sum_{j=0}^{r} \mu_{k-1}^{{(j)} \mid {(i)}} [ \hat{\mb{C}}_{k-1}^{{(j)}} +\\
           & (\hat{\mb{m}}_{k-1}^{{(j)}} - \hat{\mb{m}}_{k-1}^{(0i)})(\hat{\mb{m}}_{k-1}^{{(j)}} - \hat{\mb{m}}_{k-1}^{(0i)})\T ]
        \end{aligned}$
\EndFor
\item[]

\item[] \textbf{2) Shape state estimation with \( r \) EKFs:}
\For{\( i = 0 \) to \( r \)}
    % \item[]    Prediction Stage:
    \State \( \check{\mb{m}}_k^{(i)}=\mb{f}\left(\hat{\mb{m}}_{k-1}^{(0i)}, \mb{u}_k\right) \)
    \State \( \check{\mb{C}}_{k}^{(i)}=\mb{F}_k^{(i)} \hat{\mb{C}}_{k-1}^{(0i)}\left(\mb{F}_k^{(i)}\right)^T+\mb{Q}^{(i)} \)
    % \item[] Correction Stage:
    \State \( \mb{S}_k^{(i)} = \mb{H}_k^{(i)} \check{\mb{C}}_{k}^{(i)}\left(\mb{H}_k^{(i)}\right)\T+\mb{R}^{(i)} \)
    \State \( \mb{K}_k^{(i)}=\check{\mb{C}}_{k}^{(i)}\left(\mb{H}_k^{(i)}\right)\T\left(\mb{S}_k^{(i)}\right)^{-1} \)
    \State \( \hat{\mb{m}}_{k}^{(i)}=\check{\mb{m}}_{k}^{(i)}+\mb{K}_k^{(i)}\left(\mb{y}_k-\mb{h}_p\left(\check{\mb{m}}_{k}^{(i)}\right)\right) \)
    \State \( \hat{\mb{C}}_{k}^{(i)}=\left(\mb{I}-\mb{K}_k^{(i)} \mb{H}_k^{(i)}\right) \check{\mb{C}}_{k}^{(i)} \)
\EndFor

\item[]

\item[] \textbf{3) Model likelihood update:}
\For{\( i = 0 \) to \( r \)}
    \State \( \Lambda_k^{(i)}=\mathcal{N}\left(\mb{y}_k - \mb{h}_p\left(\check{\mb{m}}_{k}^{(i)}\right); \mb{0}, \mb{S}_k^{(i)}\right) \)
    \State \( \mu_k^{(i)} = \frac{\check{\mu}^{{(i)}}_k \Lambda_k^{(i)}}{\sum_{j=0}^r \check{\mu}^{{(j)}}_k\Lambda_k^{(j)}} \)
\EndFor

\item[]

\item[] \textbf{4) State estimate combination:}
   
\State \( \hat{\mb{m}}_{k}=\sum_{i=0}^r \mu_k^{(i)} \hat{\mb{m}}_{k}^{(i)} \)
\State \( \hat{\mb{C}}_{k} = \sum_{i=0}^r \mu_{k}^{{(i)}} \Bigl(\hat{\mb{C}}_{k}^{(i)}  + \left(\hat{\mb{m}}_{k}-\hat{\mb{m}}_{k}^{{(i)}}\right)\left(\hat{\mb{m}}_{k}-\hat{\mb{m}}_{k}^{{(i)}}\right)\T\Bigl) \)

\item[]

\item[] \textbf{5) TPM update:}
\For{\( i = 0 \) to \( r \)}
    \State \( \lambda_{i j}=\frac{\Lambda_i}{\Lambda_j}, \quad j=[0,1, \ldots, r], \quad i \neq j \)
    \State \textbf{Update $\hat{p}_{i j}$ using \eqref{eqn: dual_correction} or \eqref{eqn: single_correction}}

    \State \(p_{i j}=\frac{\hat{p}_{i j}}{\sum_{j=0}^M \hat{p}_{i j} }\)
\EndFor

\item[]
\EndFor

\end{algorithmic}
\end{algorithm}

\section{Observability Analysis } \label{ch: observerbility}
For each model in our IMM framework, an EKF is used to estimate the shape state \(\mb{m}\). Convergence of the EKF requires that the system be observable. In linear time-invariant (LTI) systems, observability is often checked via the rank of the observability matrix. In nonlinear systems, however, we adopt \emph{local weak observability}~\cite{hermann1977nonlinear}, which relies on Lie derivatives to capture how variations in the shape state map to sensor outputs.

Mahoney et al.~\cite{mahoney2016inseparable} also employ a rank-based observability analysis for continuum robots, emphasizing boundary constraints and sensor placement. While the core principle of examining how small perturbations in the shape state affect sensor measurements remains the same, our approach differs in several key ways: we use local weak observability via Lie derivatives, explicitly incorporate sensor noise through a \emph{noise-weighted observability matrix}, and evaluate observability via two dedicated indices. Moreover, we integrate this analysis into an EKF framework for real-time state estimation, in contrast to the more structurally focused analysis in~\cite{mahoney2016inseparable}.
\subsection{Noise-weighted Observability Matrix}
A nonlinear system is locally weakly observable if the observability rank condition holds in a neighborhood around a specific point in its state space. This condition is derived by constructing the observability matrix using successive Lie derivatives of the output function \(\mb{h}_p(\mb{m}_k)\). The \(0^\mathrm{th}\) order Lie derivative is simply \(\mb{h}_p(\mb{m}_k)\), while higher-order derivatives follow:
\begin{equation}
\left(\mathcal{L}_f^1 \mb{h}_p\right)\!(\mb{m}_k,\mb{u}_k)
    = \frac{\partial \mb{h}_p(\mb{m}_k)}{\partial \mb{m}_k}
      \,\mb{f}(\mb{m}_k,\mb{u}_k)
\label{eqn:1st_lie}
\end{equation}
\begin{equation}
\left(\mathcal{L}_f^j \mb{h}_p\right)\!(\mb{m}_k,\mb{u}_k)
    = \frac{\partial \bigl(\mathcal{L}_f^{j-1}\!\mb{h}_p\bigr)\!(\mb{m}_k,\mb{u}_k)}{\partial \mb{m}_k}
      \,\mb{f}(\mb{m}_k,\mb{u}_k)
\label{eqn:jth_lie}
\end{equation}
where \(\mb{f}\bigl(\mb{m}_k,\mb{u}_k\bigr)=\dot{\mb{m}}\bigl(k\,T_s\bigr)\) and \(T_s\) is the sampling time. Stacking these derivatives yields the observability matrix:
\begin{equation}
\label{eqn:general_observability_matrix}
\mathcal{O}_{\mb{m}}(\mb{m}_k, \mb{u}_k)
= \frac{\partial}{\partial \mb{m}_k}
  \begin{bmatrix}
   (\mathcal{L}_f^0 \mb{h}_p) (\mb{m}_k,\mb{u}_k) \\
   (\mathcal{L}_f^1 \mb{h}_p) (\mb{m}_k,\mb{u}_k) \\
   \vdots\\
   (\mathcal{L}_f^{n-1}\!\mb{h}_p)(\mb{m}_k,\mb{u}_k)
  \end{bmatrix}
\end{equation}
The shape states are locally observable at \(\bigl(\mb{m}_k,\mb{u}_k\bigr)\) if \(\mathcal{O}_{\mb{m}}(\mb{m}_k, \mb{u}_k)\) has full rank \(n\). However, this conventional matrix does not capture the impact of sensor noise.

To address varying noise conditions, we propose the \emph{noise-weighted observability matrix}, defined as
\begin{equation}
    \label{eqn:noise_weighted_observability_matrix}
    \mathcal{O}_{\mb{wm}} 
    = \mathcal{O}_{\mb{m}}^\top\, \mb{R}^{-1}\, \mathcal{O}_{\mb{m}}
\end{equation}
where \(\mb{R}\) is the sensor-noise covariance. This weighting accounts for noise levels directly in the observability assessment, enabling a more reliable analysis under varying measurement conditions.
\subsection{Assessment of Observability}
\label{ch: observability assessment}

\subsubsection{Inverse condition number}

In noisy, practical settings, relying solely on a rank test for local observability is often insufficient. Instead, the \emph{inverse condition number} is commonly used:
\begin{equation}
    inv\_cond(\mathcal{O}_{\mb{wm}})
    = \frac{\sigma_{\min}\bigl(\mathcal{O}_{\mb{wm}}\bigr)}{\sigma_{\max}\bigl(\mathcal{O}_{\mb{wm}}\bigr)}
\end{equation}
where \(\sigma_{\max}\) and \(\sigma_{\min}\) are the maximum and minimum singular values of \(\mathcal{O}_{\mb{wm}}\), respectively. As a matrix nears rank deficiency, its inverse condition number approaches zero; a value closer to 1 indicates stronger observability. Nonetheless, being dimensionless, it may not always discriminate well among different noise levels~\cite{nahvi1996noise}.

\subsubsection{Noise amplification index}

To overcome this limitation, we also use the \emph{noise amplification index}~\cite{nahvi1996noise}, defined as
\begin{equation}
    noise\_amp(\mathcal{O}_{\mb{wm}})
    = \frac{\sigma_{\min}^2\bigl(\mathcal{O}_{\mb{wm}}\bigr)}{\sigma_{\max}\bigl(\mathcal{O}_{\mb{wm}}\bigr)}
\end{equation}
A larger value indicates better observability in the presence of noise, offering additional insight beyond the inverse condition number. It is worth noting that the noise-weighted observability matrix and its associated indices can be computed in real time, enabling on-the-fly assessments of how well each sensor configuration observes the robot’s shape.

\section{Simulation Study}
\label{ch: simulation}

\par  We evaluate the proposed shape–estimation framework in simulation using a matched \texttt{Python} (NumPy/SciPy) environment. The study proceeds in three steps: (i) benchmark per-update latency of PCK variants under identical runtime conditions; (ii) quantify how sensing type and placement affect accuracy using a standard EKF with $M^{(2)}$; and (iii) under the best-performing arrangement, compare IMM–EKF variants to a single-model EKF. Throughout, we report pose- and state-space errors and relate performance to observability metrics.

\subsection{Computation Time}
\begin{figure}[!h]
	\centering
	\includegraphics[width=1\columnwidth]{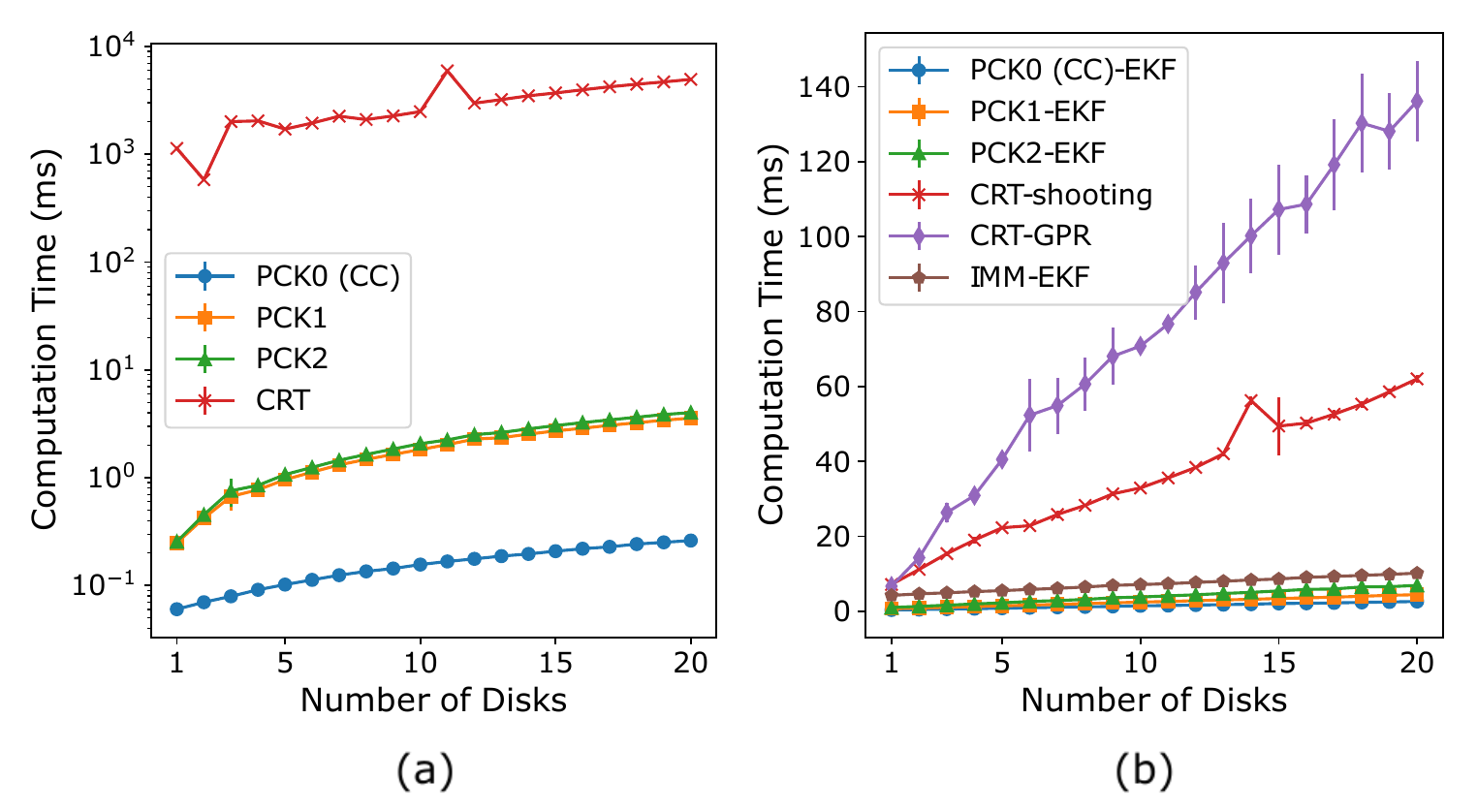}
	\caption{Algorithmic latency in a matched Python setup (single thread): (a) Forward kinematics (joint inputs $\rightarrow$ disk poses), (b) Single-update estimation (tip pose $\rightarrow$ disk poses). Values are mean ± s.d. over 10 repeated runs and are implementation-dependent.}
	\label{fig:Compute time}
\end{figure}
We benchmark wall-clock \emph{per-update latency} and scaling within a single \texttt{Python} (NumPy/SciPy) runtime on the same single-threaded machine, so methods are compared under identical interpreter/dependency conditions rather than across languages or hardware. The suite includes a CRT shooting solver and a Cosserat-prior GPR estimator adapted from public references~\cite{rao2021model,lilge2022continuum}, alongside PCK forward kinematics and the proposed IMM–PCK estimator. CRT timings are reported solely as a like-for-like runtime reference to inform loop-rate budgeting—not as an accuracy comparison or a cross-language speed ranking.

Because PCK is kinematics-only and does not require force/torque inputs, CRT was evaluated in the unloaded case (no external/actuation loads) to avoid introducing unmeasured inputs under sparse sensing and to isolate solver latency independent of load identification. All timings reflect wall-clock measurements over identical discretizations.

As shown in Fig.~\ref{fig:Compute time}, PCK variants maintained sub-10\,ms per-update latency across typical discretizations (1–20 disks) for both forward kinematics and the estimation step. IMM–PCK adds little overhead relative to single-PCK EKFs while remaining substantially faster than CRT and CRT+GPR in this matched \texttt{Python} setting. These results are specific to the shared runtime; highly optimized CRT implementations (e.g., C/C++) can be fast for many applications, but our focus is a low-latency estimator that operates without force inputs under limited sensing.
\subsection{Synthetic Data Generation}
\subsubsection{Ground truth shapes and modal coefficients}
We simulate a bending process ranging from a nearly straight configuration to a significantly bent configuration. This process is represented through a sequence of 100 samples, employing three distinct models within our model set $\mathbb{M}$. The simulations focus on the central backbone of the continuum robot, with a fixed length $L = 100\,\text{mm}$. \par
\begin{figure}[!h]
    % \vspace{5mm}
	\centering
	\includegraphics[width=0.8\columnwidth]{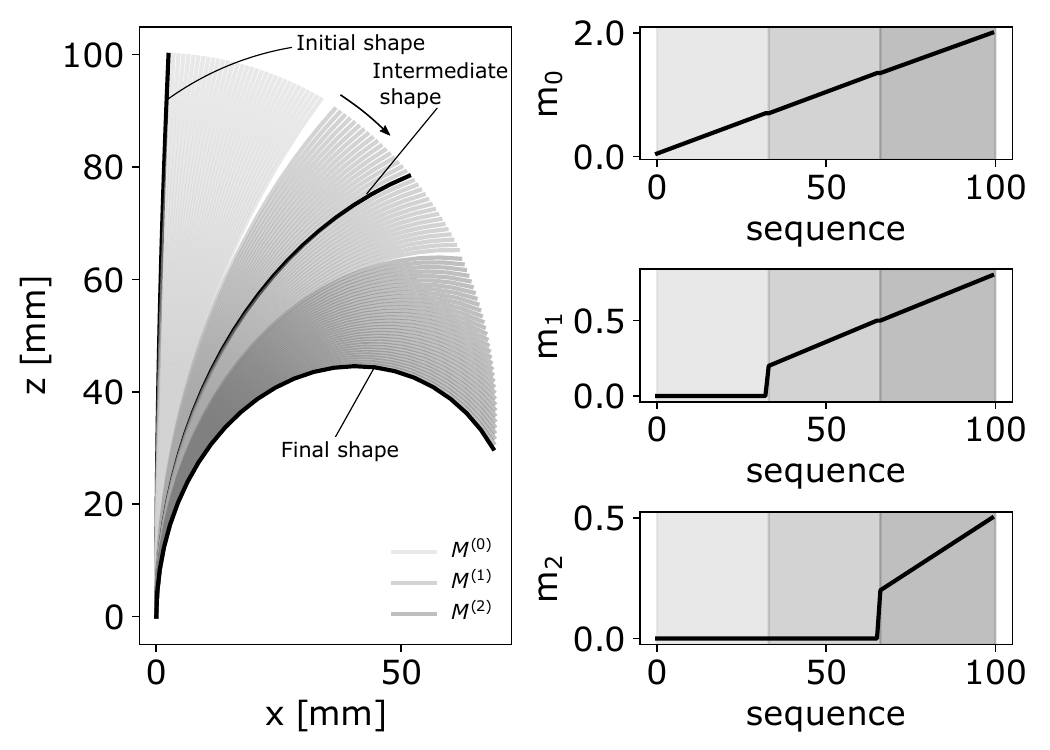}
	\caption{Simulation ground truth shapes and modal coefficients across different polynomial stages. The figure shows ground truth shapes (left) and modal coefficients $m_0$, $m_1$ and $m_2$ (right) across three stages of the simulation, each defined by a different polynomial curvature model: $M^{\mathit{(0)}}$ for Stage I, $M^{\mathit{(1)}}$  for Stage II, and $M^{\mathit{(2)}}$  for Stage III. The modal coefficients directly determine the shapes at each stage, as indicated by the shaded background corresponding to each polynomial order.}
	\label{fig: Simulated ground truth}
\end{figure}

The simulation setup corresponds to a sampling time $T_s = 0.05\, \text{s}$. The angular velocity at the tip is computed as
\begin{equation}
    \dot{\theta}_e = \frac{\theta_{e,\text{final}} - \theta_{e,\text{initial}}}{N \cdot T_s}
\end{equation}
where $N = 100$ is the number of samples. In our current setup, this yields $\dot{\theta}_e = 0.1\,\text{rad/s}$. So that the filter input at time step $k$ is given by $u_k = T_s \cdot \dot{\theta_e}$. It is important to note that this angular velocity is a design artifact chosen for this simulation. For a given motion trajectory, different sampling times (and hence different angular velocities) can be used provided that the computation time for each estimation cycle (approximately 5 ms for a standard EKF, as discussed in the preceding section) is less than $T_s$. This framework generates a comprehensive set of simulated ground truth shapes and corresponding modal coefficients, as depicted in Fig. \ref{fig: Simulated ground truth}.

\subsubsection{Measurements}
For the measurements in different sensor arrangements, we extract position and orientation at specified arc lengths based on the ground truth of modal coefficients generated previously and introduce noise to them to emulate real sensor behavior. This noise is generated by drawing random values from a zero-mean Gaussian distribution, with standard deviations set to reflect typical sensor inaccuracies encountered in practice. Specifically, we set the standard deviation for position measurement at $\mb{\nu}_\mb{r} = 1.5$ mm and for orientation measurements at $\mb{\nu}_\mb{\psi} = 1.72^{\circ}$. \par
\begin{figure}[!bh]
    \vspace{-4mm}
	\centering
	\includegraphics[width=0.8\columnwidth]{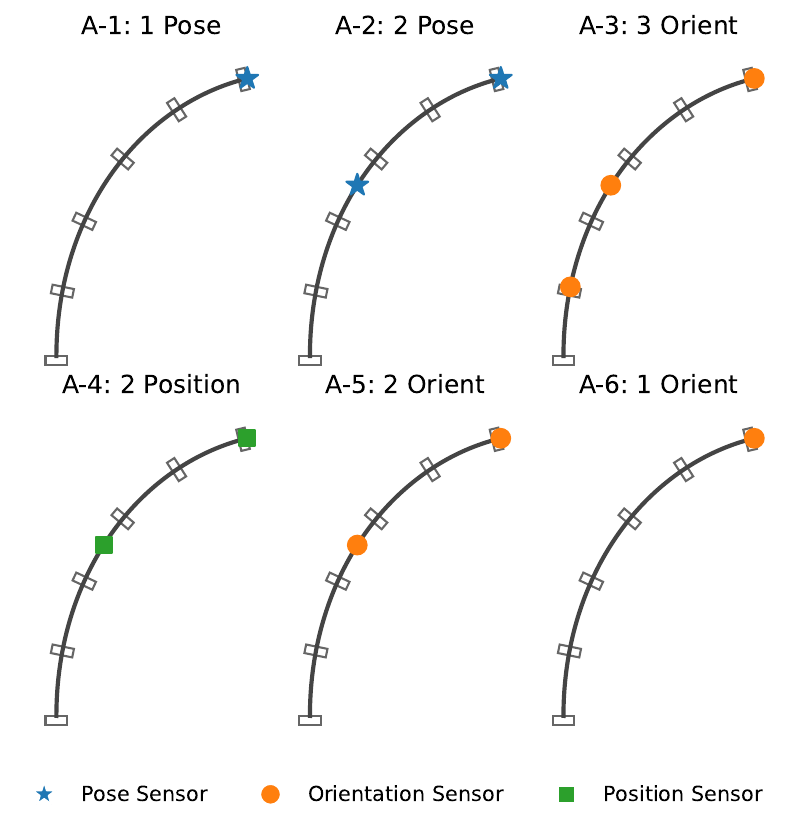}
	\caption{Sensor arrangement configurations for simulation measurements on a continuum robot. A-1 has a single pose sensor at the tip (s=1); A-2 has two pose sensors at s=1 and s=0.5; A-3 places three orientation sensors at s=1, s=0.5, and s=0.2; A-4 positions two position sensors at s=1 and s=0.5; A-5 includes two orientation sensors at s=1 and s=0.5; A-6 features a single orientation sensor at the tip.}
	\label{fig: Simulated arrangement setup}
\end{figure}
\subsection{Evaluation Metrics}
In evaluating the performance of estimators, we utilize two methodologies: error analysis in Cartesian space and error analysis in modal coefficient space. \par
In the Cartesian space, we consider two common error metrics: tip error and shape error\cite{shi2016shape}. 
At any point along the continuum robot’s arc, we define two error metrics. The position error $e_p$ is expressed as a percentage relative to the arc length up to that point, while the rotation error $e_r$ is given as the relative error in the bending angle per 100 $mm$ of arc length. Specifically, we define:
\begin{equation}
e_{p}(s) = \frac{|| p_{{est}}(s) - p_g(s) ||}{sL} \cdot 100
\label{eqn:relative_p_err}
\end{equation}
\begin{equation}
e_{r}(s) = \frac{\left| \theta_{{est}}(s) - \theta_g(s) \right|}{sL} \cdot 100
\label{eqn:relative_r_err}
\end{equation}
where $ p_{{est}}$ and $ p_g$ are the estimated and ground truth positions, $ \theta_{{est}} $ and $ \theta_g $ represent the estimated and ground truth bending angle, respectively, and $L$ is the total length of the central backbone.\par
The tip errors are given directly by \eqref{eqn:relative_p_err} and \eqref{eqn:relative_r_err} using corresponding data at the tip. The shape errors are the average errors across multiple evenly distributed locations along the arc length, including the tip, similar to the metric used in \cite{rao2021using}. Here we compute shape errors using 10 sampled locations along the arclength. \par
In modal coefficient space, we focus on the absolute error of the modal components of the shape state vector. This approach allows us to assess the accuracy of our shape representation in terms of its modal characteristics. Combining these two metrics offers a comprehensive evaluation of the estimation performance, capturing both spatial accuracy and the fidelity of the novel shape state representation using modal coefficients. \par
In addition, we investigate the observability of each sensor arrangement using indices described in section \mbox{\ref{ch: observability assessment}}.

\subsection{Sensor-Arrangement Evalutation}

\begin{figure*}[!h]
	\centering
	\includegraphics[width=0.95\columnwidth]{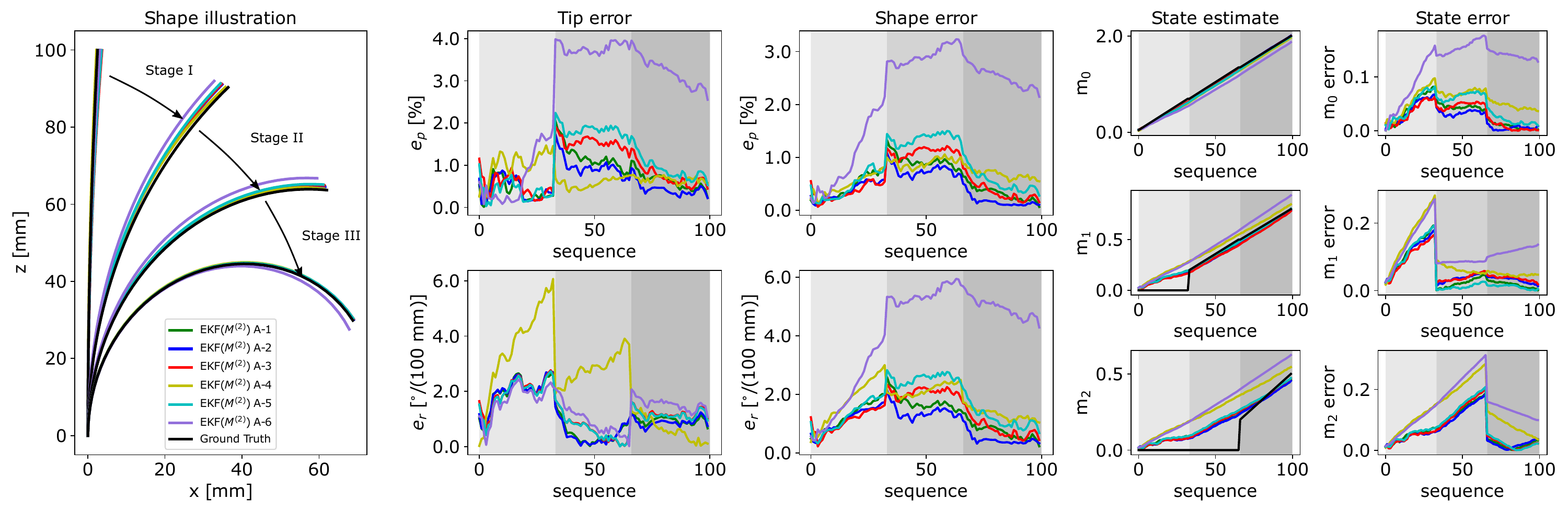}
	\caption{Estimation performance of different sensor arrangements using a standard EKF with $M^{\mathit{(2)}}$. The leftmost panel shows shape illustrations at three stages of the simulation, with the actual shape denoted as 'Ground Truth'. The subsequent panels detail the tip error, shape error, and the estimation and error of modal coefficients ($m_0$, $m_1$, $m_2$) across the sequence. Shaded areas in the error panels indicate different simulation stages defined in fig. \ref{fig: Simulated ground truth}.}
	\label{fig:Sensor Arrangement pose error}
\end{figure*}\par 

\subsubsection{Sensor setup}
In this work, we define different sensor attachment setups as 'Arrangements' (denoted as A-x, where x is the arrangement number). We investigate six representative sensor arrangements to examine how different sensor types, numbers, and placements affect shape estimation performance. Although infinitely many configurations exist, these six were chosen to reflect typical engineering constraints (e.g., weight, size, and cost) while spanning a range of design choices. Figure~\ref{fig: Simulated arrangement setup} illustrates these configurations with six example shapes.\par
Although geometric intuition can guide basic sensor placements under simple, planar bending conditions, we propose using an observability matrix to provide a more robust, nonlinear framework for more complex shapes. In particular, for continuum robots or higher-order curvature scenarios, relying on geometric intuition alone becomes impractical. The observability-based approach thus offers a systematic way to ensure that selected sensor arrangements capture enough information to accurately reconstruct the robot’s shape, even under challenging deformations.

\subsubsection{Estimator settings}
In this simulation study, we employ model $M^{\mathit{(2)}}$, characterized by a $2^{nd}$ order polynomial curvature function, in conjunction with the standard EKF to assess the influence of different sensor arrangements on shape estimation accuracy. The state vector is $\mb{m}^{\mathit{(2)}} = [m_0^{\mathit{(2)}}, m_1^{\mathit{(2)}}, m_2^{\mathit{(2)}}]\T$.\par
We initialize the state as $\mb{m}^{\mathit{(2)}} = [0, 0, 0]\T$, assuming that the continuum robot starts in a straight configuration. The standard deviation of state element is set as $\mb{\nu}_m = 0.005$. Following empirical tuning, we define the nominal process covariance matrix for model $M^{\mathit{(2)}}$ as $\mb{Q}_n = \textbf{diag}(\mb{\nu}_m^2, \mb{\nu}_m^2, \mb{\nu}_m^2)$. Similarly, the nominal measurement covariance matrix for two pose measurements is set as $\mb{R}_n = 5\hspace{0.25em}\textbf{diag}(\mb{\nu}_\mb{r}^2, \mb{\nu}_\mb{r}^2, \mb{\nu}_\mb{\psi}^2,\mb{\nu}_\mb{r}^2, \mb{\nu}_\mb{r}^2, \mb{\nu}_\mb{\psi}^2)$. From this, one can deduce the nominal measurement covariance matrices for other sensor arrangements by considering both the sensor type and the number of sensors. A detailed discussion on hyperparameter tuning and sensitivity analysis is provided in Section~\ref{ch:tuning_and_sensitivity_analysis}. \par
\subsubsection{Results}
The estimation results across six sensor arrangements in Cartesian space are illustrated in fig. \ref{fig:Sensor Arrangement pose error}. 
The right plots display the estimation results of the shape state (modal coefficients). The tip error and shape error in both position and orientation are illustrated in the middle plots. The left plot shows the ground truth and the estimated shapes at four sample points. \par
We analyze the estimation results based on the observability of each arrangement. 
When we examine the rank condition of the observability matrix for each sensor arrangement, it becomes apparent that the observability matrices for A-5 and A-6 are not full rank, indicating that these scenarios are not observable. In contrast, the other four arrangements appear to be observable. \par
As unobservable scenarios, A-5 and A-6 exhibit acceptable performance in Stage I, showing low errors in tip position and bending angles, similar to observable arrangements. However, these arrangements show distinct divergence in tip position and shape errors in Stages II and III, particularly for A-6. Despite these challenges, the bending angle errors at the tip remain consistently low across all stages for both arrangements. This stability is attributed to using orientation measurements in the estimations, which effectively maintain accuracy in the bending angle despite other increasing errors.\par
\begin{figure}[!t]
	\centering
	\includegraphics[width=0.98\columnwidth]{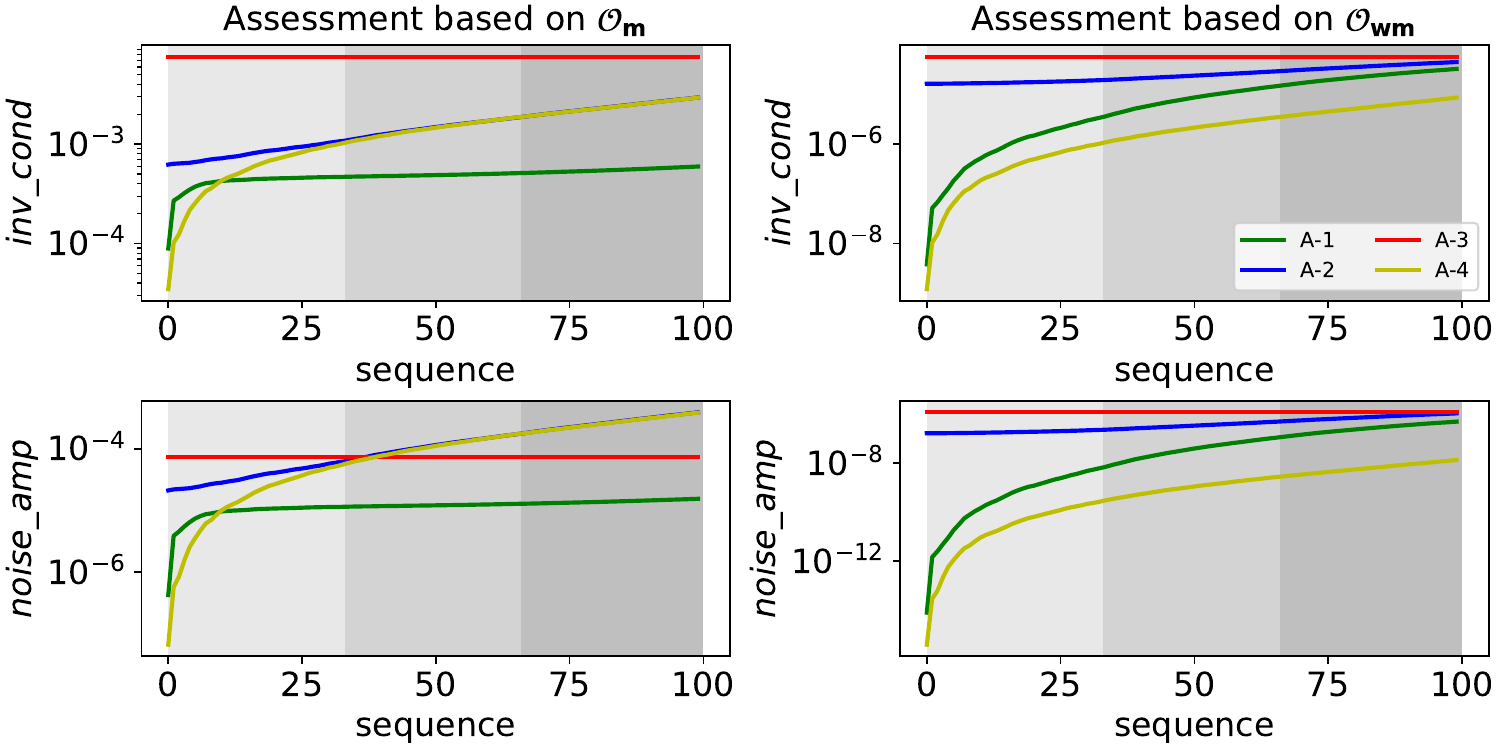}
	\caption{Observability assessment through inverse condition number and noise amplification index. The upper graph shows the inverse condition numbers, while the lower graph presents the noise amplification indices for observable arrangements. Shaded regions indicate Stages I, II, and III, respectively.}
	\label{fig:Sensor arrangement observability assessment}
\end{figure}

For the observable arrangements, most exhibit small tip and shape errors in both position and orientation across all stages, except for A-4, which utilizes two position measurements. In A-4, although position errors are relatively low, the orientation errors are considerably larger, particularly in Stages I and II. To further assess their observability, we evaluated two indices—the inverse condition number and the noise amplification index—of both the conventional observability matrix $\mathcal{O}_{\mb{m}}$ and the noise-weighted observability matrix $\mathcal{O}_{\mb{wm}}$. The results, depicted in Figure \ref{fig:Sensor arrangement observability assessment}, show conventional observability indices on the left and noise-weighted indices on the right. Unlike the indices from the conventional observability matrix, which show no correlation with the estimation results, the indices from the noise-weighted matrix demonstrate a clear correlation between observability and estimation accuracy of the modal coefficients. 
The estimation results for the shape states (modal coefficients) are consistent with the pose estimations at the tip and the overall shape. The average values of all evaluation metrics, calculated over 100 sample shapes, are presented in Table \ref{tab:estimation errors of sensor arrangement}. A-2 exhibits the best performance, with the lowest average errors.\par

\begin{table}[t]
  \caption{Sensor arrangement evaluation with standard EKF ($M^{(2)}$): mean errors}
  \label{tab:estimation errors of sensor arrangement}
  \centering
  \scriptsize
  \setlength{\tabcolsep}{3pt}
  \renewcommand{\arraystretch}{1.02}
  \begin{tabular}{l *{6}{S[table-format=2.2]}}
    \toprule
    & {A-1} & {A-2} & {A-3} & {A-4} & {A-5} & {A-6} \\
    \midrule
    Pos. [\%]—Tip                   & 0.74 & \bfseries 0.61 & 0.93 & 0.75 & 1.05 & 2.58 \\
    Pos. [\%]—Shape                 & 0.56 & \bfseries 0.42 & 0.60 & 0.69 & 0.78 & 2.15 \\
    Ori. [$^\circ$/100\,mm]—Tip     & 1.13 & \bfseries 1.10 & 1.26 & 2.34 & 1.24 & 1.38 \\
    Ori. [$^\circ$/100\,mm]—Shape   & 1.23 & \bfseries 0.99 & 1.33 & 1.74 & 1.66 & 4.14 \\
    State [$\times 10^{-2}$]—$m_0$  & 3.16 & \bfseries 2.51 & 2.91 & 5.64 & 4.27 & 12.47 \\
    State [$\times 10^{-2}$]—$m_1$  & 5.11 & 5.88 & 5.76 & 8.95 & \bfseries 4.54 & 11.70 \\
    State [$\times 10^{-2}$]—$m_2$  & 8.08 & \bfseries 7.58 & 8.04 & 15.72 & 8.74 & 18.33 \\
    \bottomrule
  \end{tabular}
\end{table}
\subsection{IMM-EKF Evaluation}
This section compares the performances of IMM-EKF algorithms - DAIMM-EKF and SAIMM-EKF - against the conventional IMM-EKF (CIMM-EKF)  and the standard EKF with $M^{\mathit{(2)}}$ under A-2.
\subsubsection{Estimator settings}
We use the same EKF settings of sensor arrangement evaluation for elementary EKFs of the IMM-EKF evaluation.
As for IMM, we initialize TPM as follows, assuming no prior knowledge:\par
\begin{equation}
\mb{P}=\left[\begin{array}{ccc}
0.7 & 0.15 & 0.15 \\
0.15 & 0.7 & 0.15 \\
0.15 & 0.15 & 0.7
\end{array}\right]
\end{equation}
The initialized model probability is set as
\begin{equation}
\mu_0^i=\frac{1}{3}, \quad i=[0,1, \ldots, r]
\end{equation} \par
Specifically, one more step needs to be considered for the DAIMM-EKF - setting up the likelihood ratio threshold $\mathrm{Th}_{i}$. Through our empirical tuning, the estimator yields enhanced performance when setting thresholds as follows $\mathrm{Th}_{0} = 1.2$, $\mathrm{Th}_{1} = 0.75$, $\mathrm{Th}_{2} = 0.75$.

\subsubsection{Results}
\begin{figure*}[!h]
	\centering
	\includegraphics[width=0.95\columnwidth]{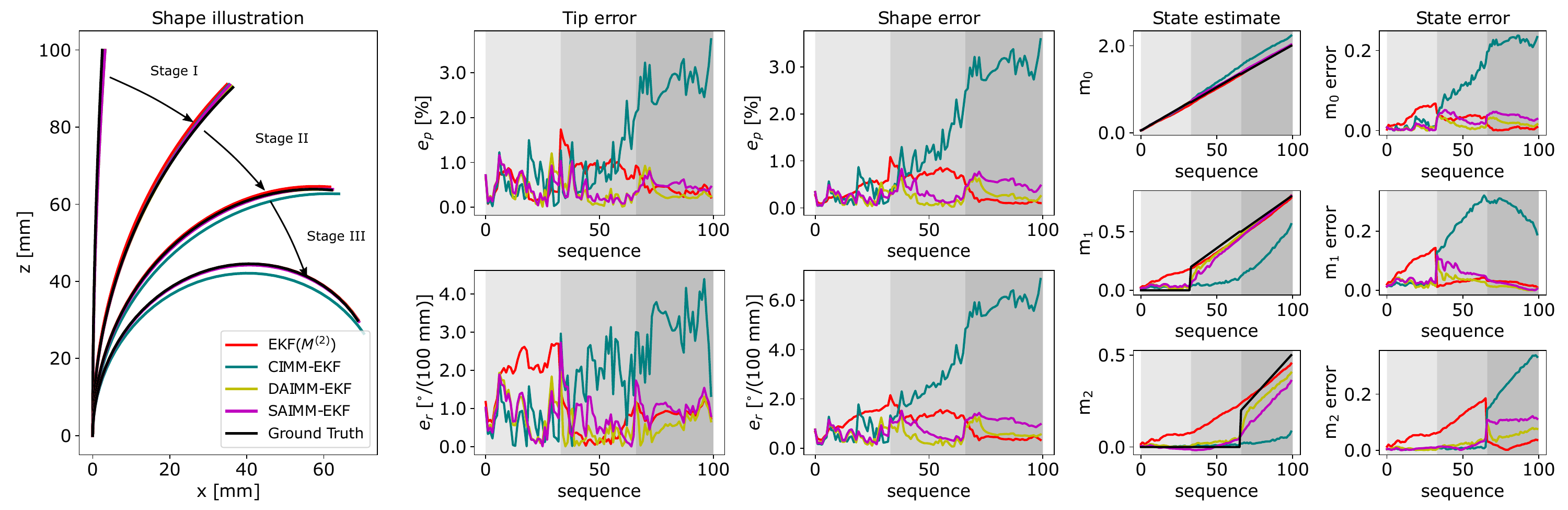}
	\caption{Comparative performance of DAIMM-EKF, SAIMM-EKF, CIMM-EKF and elementary EKF with $M^{\mathit{(2)}}$ under sensor arrangement 2. The left panel illustrates the robot's shape at key sample points against the ground truth. Middle panels quantify tip and shape errors in position and orientation. Right panels display estimation results and their errors for modal coefficients $m_0$, $m_1$, $m_2$. }
	\label{fig:IMM pose error}
\end{figure*}\par 
The pose errors for the four estimators are displayed in Fig. \ref{fig:IMM pose error}. Again, shape errors follow the trends observed in tip errors but with reduced magnitudes. The standard EKF exhibits a smoother error profile compared to the IMM variants. In Stage I, IMM estimators surpass the elementary EKF in accuracy, particularly evident in bending angle errors at the tip. Upon transitioning to a more complex model, CIMM-EKF's performance dips due to its unchanging TPM and dimensional bias issues. In contrast, DAIMM-EKF and SAIMM-EKF demonstrate superior error reduction, outperforming the standard EKF. \par
The right plots of Fig. \ref{fig:IMM pose error} illustrated the errors of modal coefficients, clearly highlighting the distinctions between the elementary EKF and IMM-EKF estimators. The elementary EKF excels when its model aligns with the true system model, showing optimal estimation accuracy. However, its performance significantly declines in scenarios with a mismatch between the utilized and true models. Conversely, the IMM-EKF estimators consistently yield better results when dealing with true models of smaller dimensionality. Across all three IMM estimators, the estimation of $m_0$ is comparably effective. Yet, for $m_1$ and $m_2$, DAIMM-EKF outperforms SAIMM-EKF, while CIMM-EKF lags, exhibiting the least favorable estimation accuracy.\par

\begin{table}[t]
  \caption{MM-EKF Evaluation: mean errors}
  \label{tab:imm average error}
  \centering
  \scriptsize
  \setlength{\tabcolsep}{3pt}
  \renewcommand{\arraystretch}{1.02}
  \begin{tabular}{l S[table-format=2.2] S[table-format=2.2] S[table-format=2.2] S[table-format=2.2]}
    \toprule
    & {EKF $M^{(2)}$} & {CIMM} & {DAIMM} & {SAIMM} \\
    \midrule
    Pos. [\%]—Tip                & 0.61 & 1.41 & \bfseries 0.38 & 0.43 \\
    Pos. [\%]—Shape              & 0.42 & 1.49 & \bfseries 0.24 & 0.36 \\
    Ori. [$^\circ$/100\,mm]—Tip  & 1.10 & 1.90 & \bfseries 0.63 & 0.87 \\
    Ori. [$^\circ$/100\,mm]—Shape& 0.99 & 2.98 & \bfseries 0.54 & 0.80 \\
    State [$\times 10^{-2}$]—$m_0$ & 2.51 & 11.46 & \bfseries 1.64 & 2.63 \\
    State [$\times 10^{-2}$]—$m_1$ & 5.88 & 21.98 & \bfseries 3.90 & 5.51 \\
    State [$\times 10^{-2}$]—$m_2$ & 7.58 & 11.15 & \bfseries 3.76 & 5.18 \\
    \bottomrule
  \end{tabular}
\end{table}

The average errors for all estimation metrics are presented in Table. \ref{tab:imm average error}. DAIMM-EKF demonstrates superior performance, recording the lowest average errors: 0.38 [\%] for position tip, 0.24 [\%] for shape error in position, 0.63 [$^\circ$/100 mm] for bending angle at the tip, and 0.54 [$^\circ$/100 mm] for shape error in bending angle. In terms of qualitative performance, both DAIMM-EKF and SAIMM-EKF surpass the standard EKF in accuracy, while CIMM-EKF falls short. \par
\begin{figure}[!t]
	\centering
	\includegraphics[width=1\columnwidth]{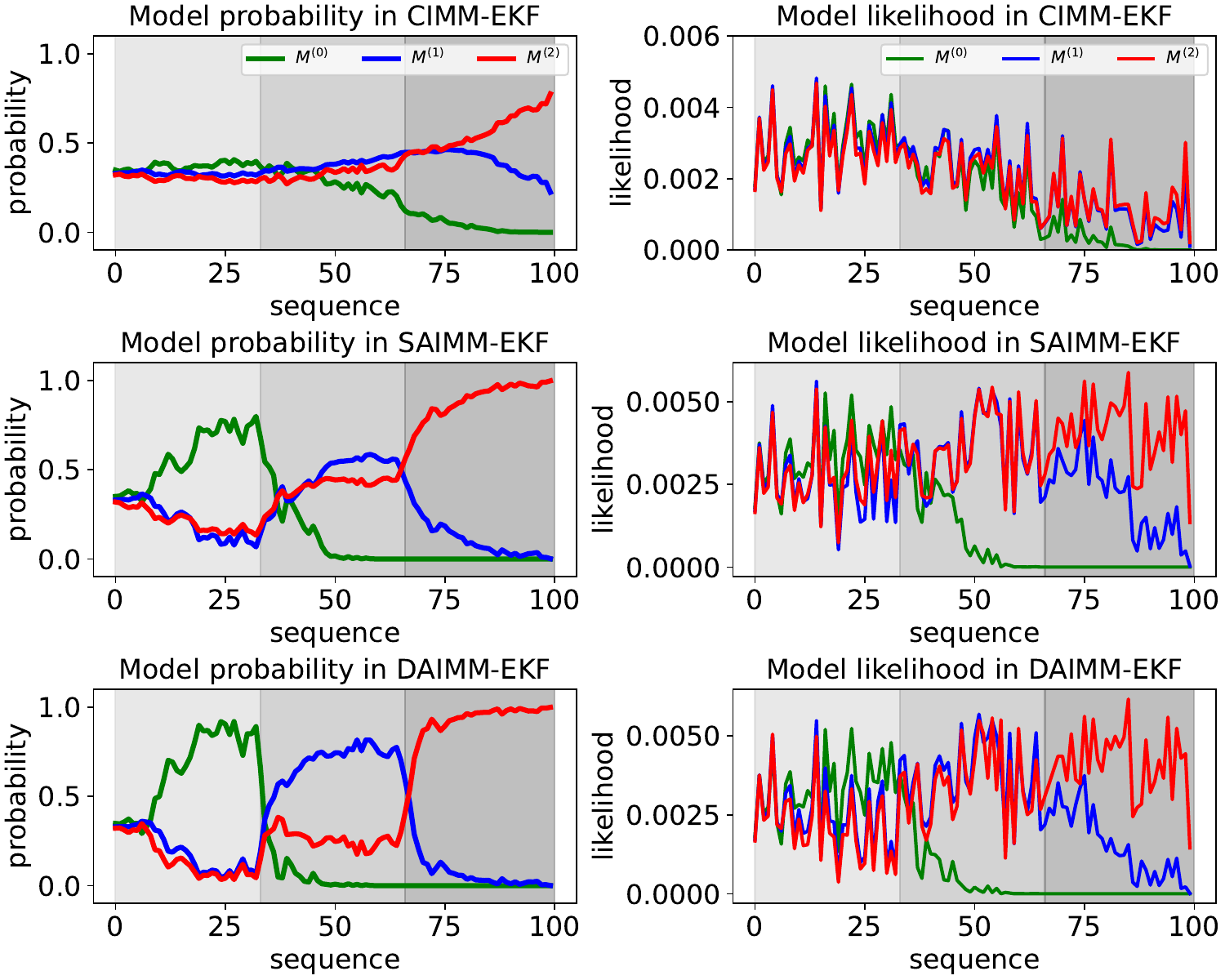}
	\caption{Model probabilities and likelihoods of IMM-EKF estimators. Each panel shows the probabilities/likelihoods for $M^{\mathit{(0)}}$ (green), $M^{\mathit{(1)}}$ (blue), and $M^{\mathit{(2)}}$(red) across 100 sequences, highlighting how each estimator adapts its model preference over time.}
	\label{fig:IMM_probability_and_likelihood}
\end{figure}
The left panels in Fig.~\ref{fig:IMM_probability_and_likelihood} illustrate the model probabilities for CIMM, DAIMM, and SAIMM, aligned with the sequential use of $M^{\mathit{(0)}}$, $M^{\mathit{(1)}}$, and $M^{\mathit{(2)}}$ in generating the ground truth data. A higher model probability for $M^{(i)}$ indicates closer alignment with the true model, enhancing IMM estimation accuracy. The trendline for CIMM exhibits slow adjustments to model transitions, reflecting the delayed response to changes in the true model.\par
To address this limitation, DAIMM introduces dual correction functions targeting model probability and likelihood ratio alongside a straightforward approach to mitigating dimensional bias. SAIMM, on the other hand, employs a single correction function focused on model probability and a more intricate method for dimensional bias correction. Both DAIMM and SAIMM demonstrate quicker adaptation to model transitions, with DAIMM responding more rapidly than SAIMM, significantly reducing lag errors. This rapid response capability contributes to DAIMM's superior estimation performance. \par
The right panels in Fig.~\ref{fig:IMM_probability_and_likelihood} highlight how the likelihood of each model in CIMM, DAIMM and SAIMM evolves over time. A key challenge with DAIMM is fine-tuning the threshold for the minimum likelihood ratio, which triggers the second correction function. Initially, when $M^{\mathit{(0)}}$ is the true model, all three models have similar likelihoods, with $M^{\mathit{(0)}}$ slightly higher. When the model transitions to $M^{\mathit{(1)}}$, the likelihood of $M^{\mathit{(0)}}$ drops sharply, while $M^{\mathit{(1)}}$ and $M^{\mathit{(2)}}$ have comparable likelihoods. If $M^{\mathit{(2)}}$ is the true model, its likelihood clearly exceeds that of the other models. However, distinguishing between models is difficult when the true model has lower dimensionality, due to similar likelihoods. To address this, we use model-specific thresholds for the minimum likelihood ratio, customized for each model's characteristics.
\subsection{Sensitivity to Noise Covariances}
\label{ch:tuning_and_sensitivity_analysis}  
\begin{figure}[!t]
	\centering
    \includegraphics[width=1\columnwidth]{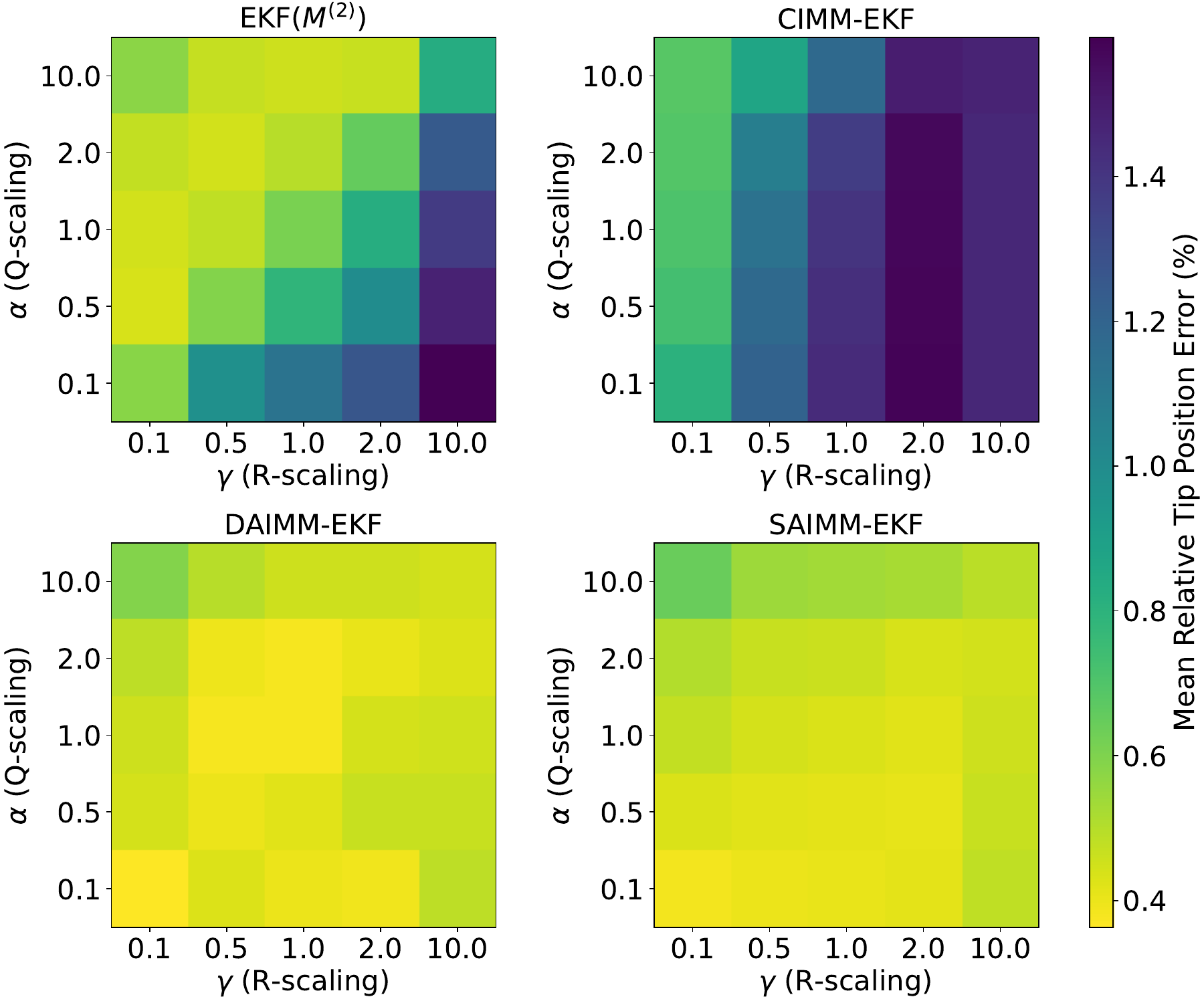}
	\caption{Sensitivity of EKF($M^{\mathit{(2)}}$), CIMM-EKF, DAIMM-EKF and SAIMM-EKF to different process and measurement noise levels. The horizontal axis $(\gamma)$ scales the nominal measurement noise covariance $\mb{R}_{n}$, and the vertical axis ($\alpha$) scales the nominal process noise covariance $\mb{Q}_{n}$. Each cell's color represents the mean relative tip position error (in $\%$ ) obtained over 25 case studies, with warmer (yellow) indicating lower error and cooler (blue) indicating higher error.}
\label{fig:hyperparameter_tuning_heatmaps}
\end{figure}

Hyperparameter tuning for state estimators is non-trivial, as their performance depends critically on the choice of process and measurement noise covariances. To evaluate robustness to mis-tuning, we conducted a sensitivity analysis on a standard EKF with $M^{\mathit{(2)}}$, a CIMM-EKF, and two enhanced IMM-EKF variants (DAIMM-EKF and SAIMM-EKF). Figure~\ref{fig:hyperparameter_tuning_heatmaps} presents heatmaps of the mean relative tip position error for each estimator, where the nominal process noise covariance $\mb{Q}_n$ and measurement noise covariance $\mb{R}_n$ are scaled by factors in $\{0.1,0.5,1,2,10\}$, yielding 25 distinct configurations. \par
In the top-left panel (EKF($M^{\mathit{(2)}}$)), warm colors (yellow) indicate lower errors and cool colors (blue) indicate higher errors, illustrating how sensitive the single-model EKF can be to noise mis-tuning. The remaining panels show the IMM-based variants (CIMM, DAIMM, and SAIMM), where, except for CIMM, the IMM approaches maintain relatively low error over a broader range of scaling factors. These results suggest that DAIMM-EKF and SAIMM-EKF, in particular, offer improved robustness to process and measurement noise covariances compared to the single-model EKF.

\section{Experimental Validation} \label{ch: experiment}
\begin{figure}[!h]
	\centering
	\includegraphics[width=0.8\columnwidth]{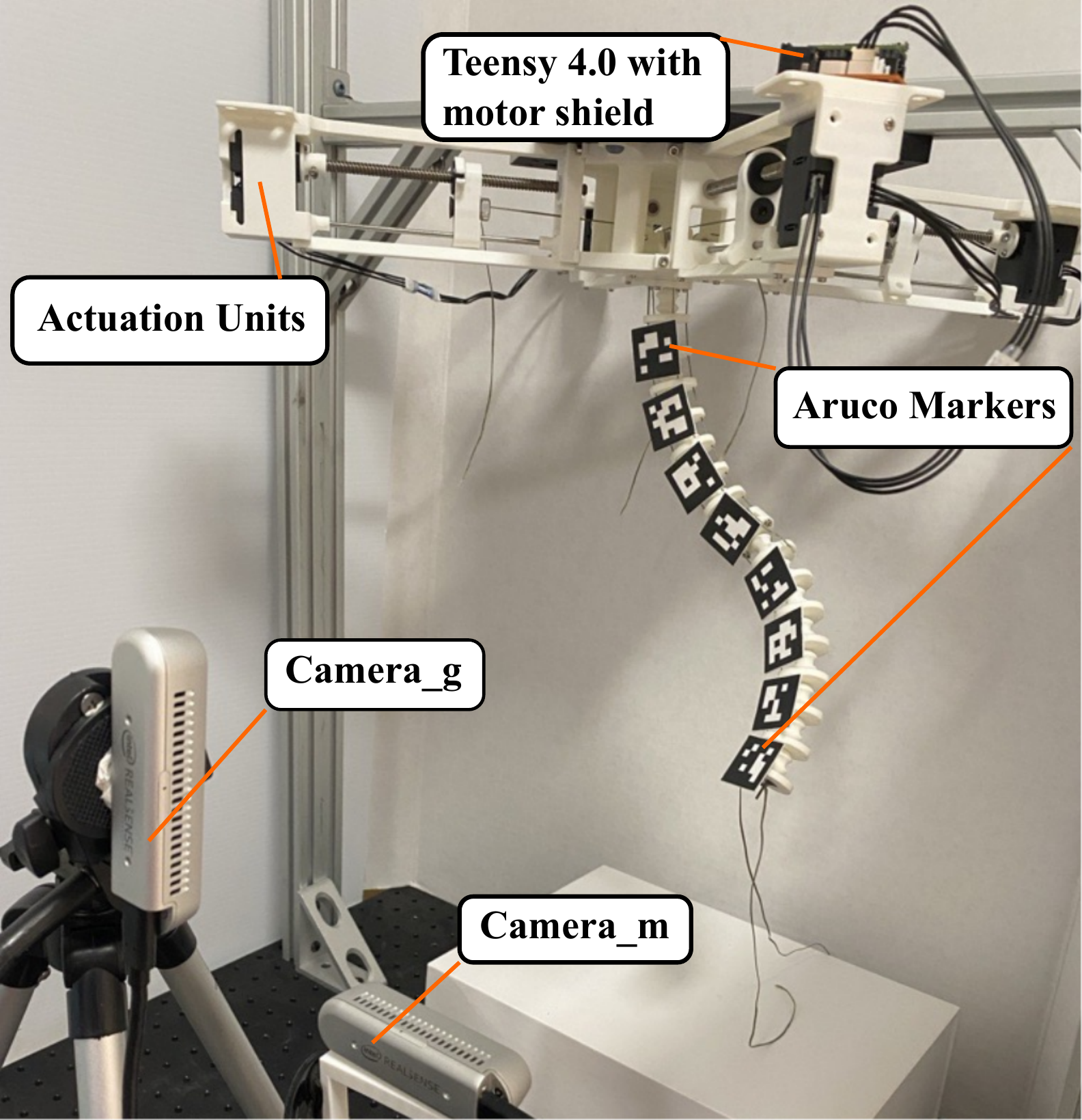}
	\caption{Experimental setup including the cable-driven continuum robot prototype, Aruco markers, and two cameras}
	\label{fig:Experimental Setup}
\end{figure}
In this section, we validate our estimation algorithms experimentally using the developed cable-driven continuum robot prototype, as shown in Fig.~\ref{fig:Experimental Setup}. To expand the workspace and enable more varied loading scenarios, we developed a two-segment design, with each segment independently actuated by routed cables. This allows complex bending profiles and continuous loads during motion, effectively demonstrating real-time estimation under nontrivial trajectories and varying loads.
The continuum instrument prototype consists of a central backbone constructed from super-elastic NiTi rod (diameter: 1.524 mm) and multiple 3D-printed spacer disks guiding four actuation wires. Two prototypes were fabricated: (1) a single-segment instrument with a length of 375 mm, and (2) a two-segment instrument, each segment having a length of 160 mm.
Actuation is achieved using lead-screw mechanism modules driven by servo motors (Dynamixel XL-430, Robotis Inc.). Control commands are managed by a Teensy 4.0 board microcontroller, which is paired with a customized drive shield.
The robot design accommodates up to four pulling wires for actuation, each connected to an individual actuation module.
\subsection{Sensor Arrangement}
To accurately reconstruct the ground truth shape of the continuum robot during bending, we employed two stereo cameras—$\texttt{camera\_g}$ (Intel RealSense D415) and $\texttt{camera\_m}$ (Intel RealSense D435i)—in conjunction with ArUco markers (shown in Fig. \ref{fig:Experimental Setup}) to capture poses along the robot's arc length. Both cameras operate at a sampling rate of 30 Hz, with $\texttt{camera\_g}$ (D415) primarily capturing the ground truth by detecting all markers and $\texttt{camera\_m}$ (D435i) measuring the pose specifically at the instrument's tip. \par
To ensure accurate alignment between measurements and ground truth data, we conducted extrinsic parameter calibration between the two cameras using a ChArUco calibration board. Twenty calibration images were captured to compute a static transformation matrix between camera frames. After calibration, the average position error and orientation error between measurement and ground truth were approximately 4 mm and 3 degrees, respectively. \par
ArUco markers were affixed to the front of each disk along the robot, with each marker's transformation to the corresponding frames \{Gs\} predefined. Additionally, an ArUco marker was placed at the robot's base prior to testing, enabling accurate computation of the transformation from the camera coordinate frame to the robot's base frame. \par
Although our previous simulation study showed that multiple sensors (e.g., a tip and a middle marker) can provide better absolute accuracy, here we purposely opt for a minimal single-sensor arrangement in order to focus on comparing the proposed IMM-EKF variants with conventional single-EKF methods. This helps ensure that any observed performance gains arise from the estimator itself rather than from having additional measurement data. Nevertheless, using only one sensor does not obscure the estimator's comparative performance. In fact, employing multiple sensors generally improves any estimation approach, so our choice here highlights the robustness and effectiveness of the proposed methods even in a sparse measurement scenario.
Using just this single sensor arrangement, we aim to provide a focused comparison with conventional estimators, highlighting the effectiveness of our methods under practical conditions.

\subsection{Data Collection and Processing}
The robot bending experiments were conducted with each actuation wire driven at a linear velocity of 0.02 inch/s, resulting in an approximate bending angle change of 3.6 degrees per second. This speed was carefully chosen to ensure reliable ArUco marker detection given practical constraints from the camera hardware and lighting conditions.
\begin{figure}[!h]
	\centering
    \includegraphics[width=1\columnwidth]{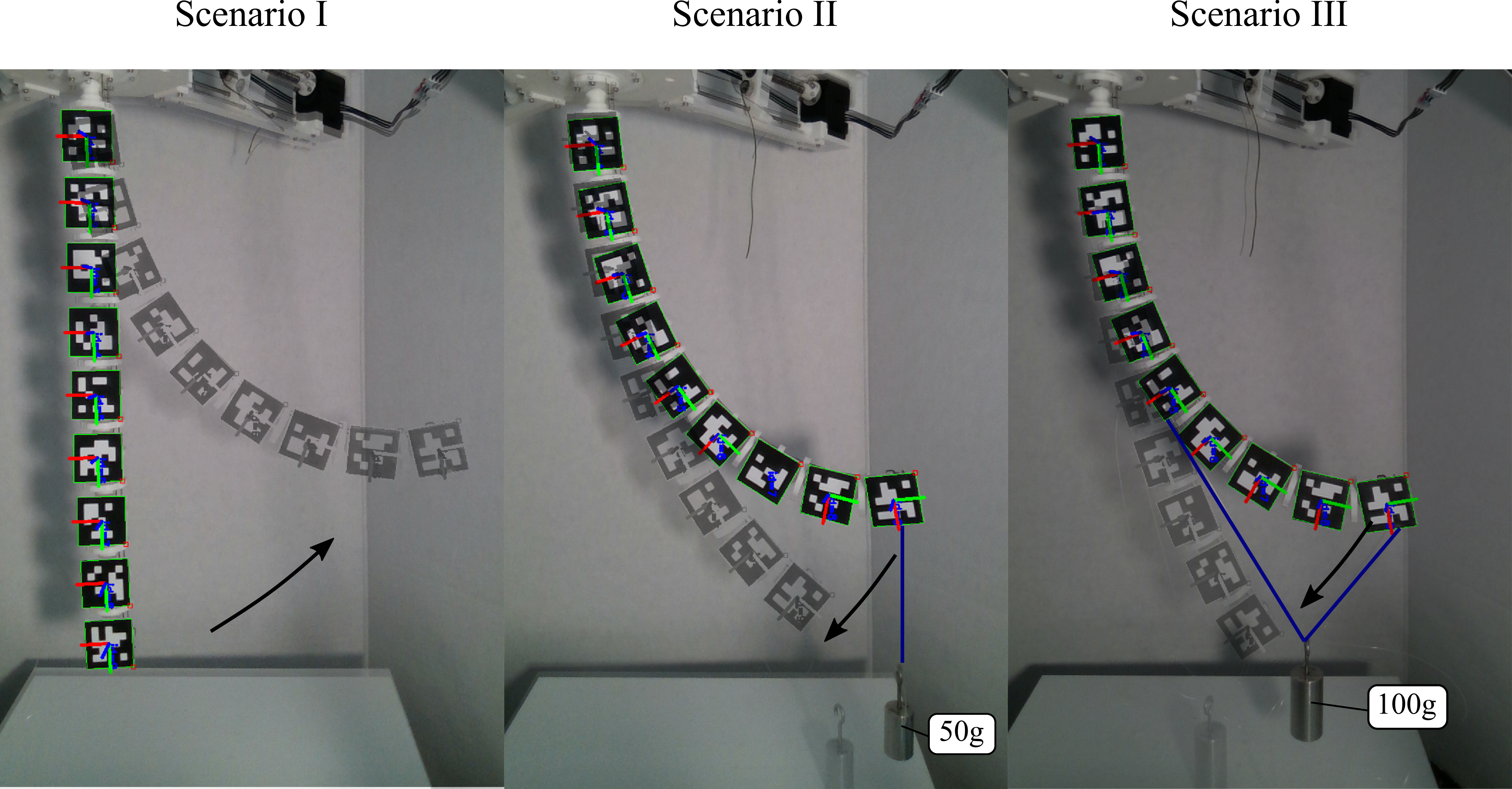}
	\caption{Single-Segment bending scenarios: (I) Straight to large bending, (II) Tip-loaded (50g) unloading, and (III) Distributed-Load (100g) unloading}
	\label{fig: One-segment bending scenarios}
\end{figure}

The experimental validation employed four distinct scenarios, comprising three single-segment bending cases (Fig. \ref{fig: One-segment bending scenarios}) and one two-segment bending case (left panel of Fig. \ref{fig:IMM exp pose est result_2seg}). Scenario I: free bending from an initially straight configuration (no external load). Scenario II: tip unloading—the tip initially carries a 50 g mass that is placed onto a fixed platform during the motion. Scenario III: distributed loading—a wire threaded through the last ten disks is pulled downward by a 100 g mass. Scenario IV: two-segment bending with independent tip loads—100 g at the first-segment tip and 50 g at the second-segment tip.
All force loading(unloading) was performed by manually adding(removing) weights in a quasi-static manner. Each loading(unloading) cycle included a brief ramp-up(down) (around 2 s), followed by a constant-load phase of a few seconds corresponding to the continuum robot motion time. Following our simulation methods, changes in the bending angle were converted into motor commands to actuate the wire consistently using inverse kinematics of \textit{joint-to-configuration} space, as detailed in the supplementary Appendix A. During each scenario, we collected raw pose data from all Aruco markers throughout the bending process within a ROS2 environment\cite{doi:10.1126/scirobotics.abm6074}, enabling us to compute disk poses relative to the base frame.
Since our study focuses solely on in-plane bending, determining the bending plane was essential. We collected all y-axes from the Frames \{Gs\} on the disks and employed Singular Value Decomposition (SVD) to pinpoint the principal y-axis. This axis helped establish the normal to the bending plane. Using this principal y-axis alongside the origin of the base frame, we defined the bending plane, enabling the projection of the 6 DOF pose data into 3 DOF within this plane. The projected poses from the Aruco marker at the tip detected by $\texttt{camera\_m}$ served as the measurement data for our analyses. 
\begin{figure*}[!h]
	\centering
	\includegraphics[width=1\columnwidth]{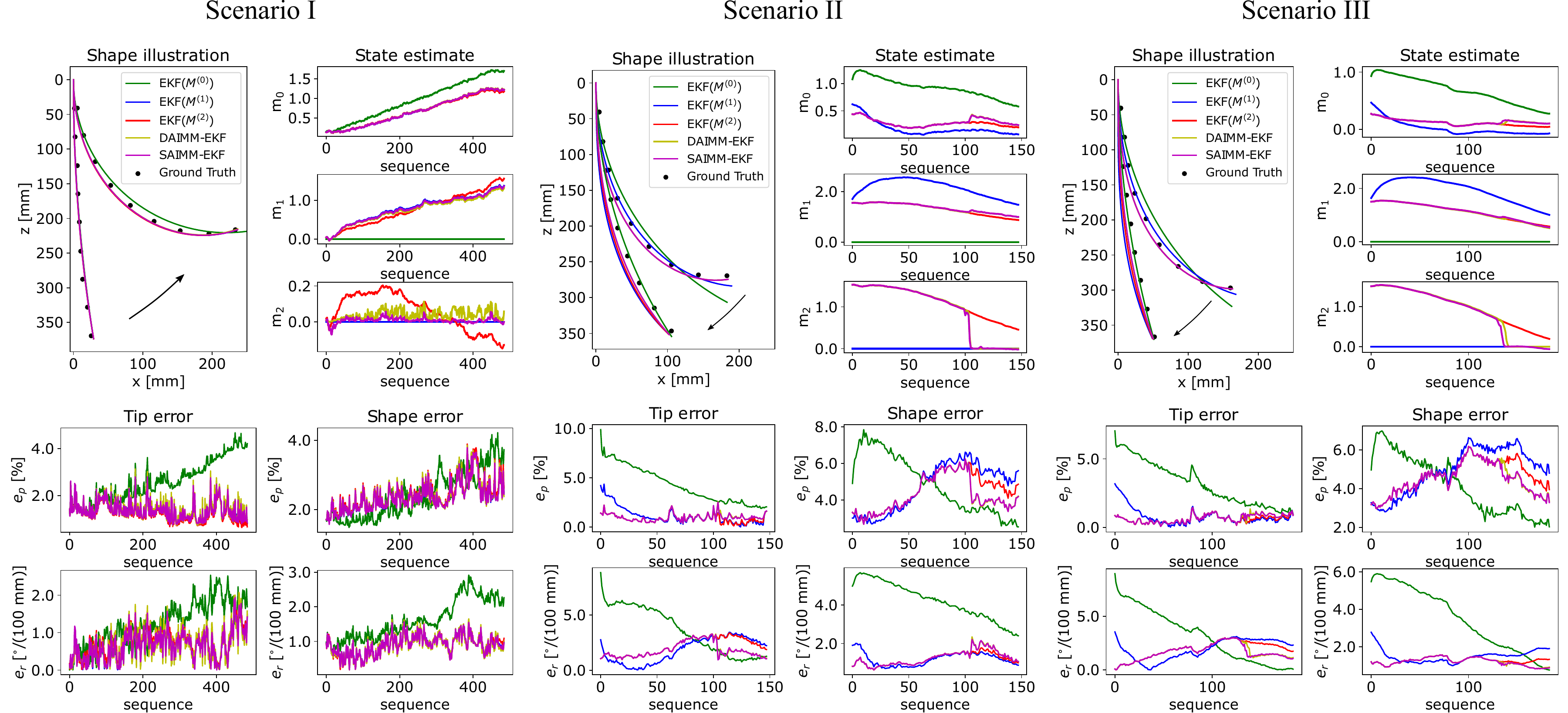}
	\caption{Comparative evaluation of single EKFs (Constant curvature(CC) $M^{\mathit{(0)}}$, 1st-order PCK $M^{\mathit{(1)}}$, 2nd-order PCK $M^{\mathit{(2)}}$,) and IMM-EKF variants (SAIMM-EKF and DAIMM-EKF) in Single-segment continuum robot bending scenarios.}
	\label{fig:IMM exp pose est result}
\end{figure*}\par

\subsection{Estimator Settings}
In the experimental scenarios involving the four bending processes of the continuum robot prototype, we deployed three elementary EKFs with $M^{\mathit{(0)}}$, $M^{\mathit{(1)}}$ and $M^{\mathit{(2)}}$, alongside two IMM-EKF variants—DAIMM-EKF and SAIMM-EKF. It was observed that $M^{\mathit{(0)}}$ did not fit well with complex loading configurations, leading to its exclusion from the IMM interaction analysis. Consequently, the model set was redefined as $\mathbb{M} = \left\{M^{\mathit{(1)}}, M^{\mathit{(2)}}\right\}$ for evaluating IMM interaction in these tests. The parameter settings for these models were consistent with those used in simulations.
\subsection{Results and Discussion}
\begin{figure*}[!h]
	\centering
	\includegraphics[width=1\columnwidth]{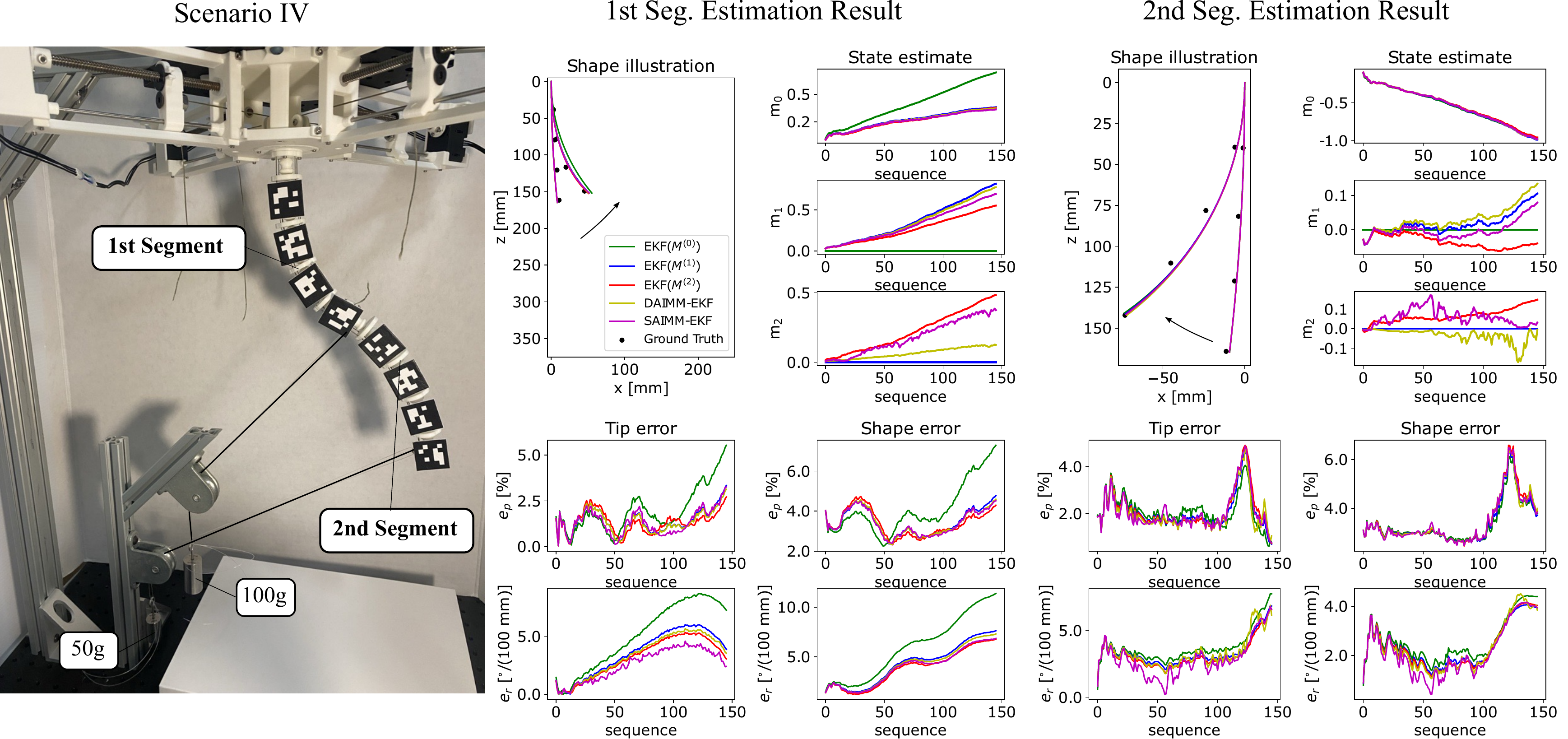}
	\caption{Comparative evaluation of single EKFs (Constant curvature(CC) $M^{\mathit{(0)}}$, 1st-order PCK $M^{\mathit{(1)}}$, 2nd-order PCK $M^{\mathit{(2)}}$,) and IMM-EKF variants (SAIMM-EKF and DAIMM-EKF) in Two-segment continuum robot bending scenario with external loading.}
	\label{fig:IMM exp pose est result_2seg}
\end{figure*}

The pose and shape state estimation outcomes from the four experimental scenarios are comprehensively illustrated in Figs. \ref{fig:IMM exp pose est result} and \ref{fig:IMM exp pose est result_2seg}, with quantitative results summarized in Table~\ref{tab:exp_error}. Overall, the proposed IMM-EKF variants—SAIMM-EKF and DAIMM-EKF—consistently outperform or closely match the performance of standard single-model EKFs (Constant Curvature $M^{\mathit{(0)}}$
 , 1st-order PCK $M^{\mathit{(1)}}$
 , and 2nd-order PCK $M^{\mathit{(2)}}$ across diverse bending and loading conditions.
All methods operate under the same limited-sensing input (single tip pose); baselines include both the classical CC EKF and single-PCK EKFs using the same polynomial orders as the IMM candidates to attribute gains to multi-model adaptation rather than a change of parameterization.

Specifically, SAIMM-EKF exhibits robust performance, attaining the lowest or near-lowest errors across most scenarios and metrics. In Scenario~I (free bending), SAIMM-EKF surpasses single-model EKFs; in Scenario~II (tip-load), it reduces tip-position mean error from 4.24\% (CC) to 0.90\%, comparable to PCK-2 (0.88\%). Although the single-model PCK-2 EKF occasionally yields the smallest error, its performance is less uniform across scenarios (e.g., higher shape errors in Scenario~I). Overall, SAIMM-EKF provides stable accuracy across regimes, supporting practical use when operating conditions vary.

DAIMM-EKF also performs favorably, often matching or slightly surpassing EKFs with PCK. Its performance suggests an effective but occasionally aggressive adaptation strategy, resulting in good overall accuracy but somewhat reduced robustness compared to SAIMM-EKF.

The two-segment bending experiment (Scenario IV) further validates the proposed IMM-EKFs' advantages. SAIMM-EKF particularly excels in rotation estimation accuracy for both segments, considerably outperforming the other estimators. This outcome indicates its superior capability to handle complex multi-segment interactions and external load-induced dynamics. In terms of position accuracy, differences among estimators in Scenario IV remain minor, underscoring rotation estimation as the more critical differentiator among methods in multi-segment scenarios.

Although direct verification of modal coefficient estimates against a known ground truth is impractical, the observed modal coefficient trajectories from SAIMM-EKF and DAIMM-EKF typically fall between the estimations from the 1st-order and 2nd-order polynomial models. Given the consistently strong estimation performance of SAIMM-EKF, its modal coefficient estimates likely provide a closer representation of the true underlying system states.

Finally, with respect to existing literature, direct quantitative comparisons are hindered by differing experimental conditions (e.g., sensor noise, hardware design, or force profiles) that heavily affect estimation outcomes. Nonetheless, the Constant-Curvature EKF ($M^{\mathit{(0)}}$) remains a widely accepted baseline\cite{ataka2016real, loo2019non, peng2024tendon}, and 1st-order polynomial curvature ($M^{\mathit{(1)}}$) aligns with popular linear-curvature assumptions (e.g., Euler curves\cite{rao2022shape} or Clothoid curves\cite{gandhi2023shape}). Hence, assessing our IMM-EKFs under identical conditions against these standard single-model EKFs demonstrates the measurable advantages and robustness offered by the proposed stochastic observer in polynomial curvature space. The ability to flexibly employ higher-order models ($M^{\mathit{(2)}}$) further underscores its potential for improved accuracy when the application demands richer shape representations.

\begin{table}[h!]
\centering
\scriptsize
\caption{Estimation Errors of the Experiments}
\begin{tabular}{llrrrrr}
\hline
\textbf{Scenario} & \textbf{Metric} & \textbf{EKF(0)} & \textbf{EKF(1)} & \textbf{EKF(2)} & \textbf{DAIMM} & \textbf{SAIMM} \\
\hline
\multirow{4}{*}{Scen.-I}
 & P-T & \underline{2.62} & 1.34 & 1.42 & 1.35 & \textbf{1.30} \\
 & P-S & 2.40 & \textbf{2.36} & \underline{3.43} & 2.37 & \textbf{2.36} \\
 & R-T & \underline{1.22} & 0.66 & 0.76 & 0.66 & \textbf{0.65} \\
 & R-S & \underline{1.61} & 0.90 & 0.90 & \textbf{0.88} & 0.89 \\
\hline
\multirow{4}{*}{Scen.-II}
 & P-T & \underline{4.24} & 1.15 & \textbf{0.88} & 0.98 & 0.90 \\
 & P-S & \underline{4.88} & 4.73 & 4.53 & \textbf{4.22} & \textbf{4.22} \\
 & R-T & \underline{3.62} & 1.84 & 2.19 & 1.82 & \textbf{1.81} \\
 & R-S & \underline{4.23} & 1.22 & \textbf{1.20} & 1.25 & 1.25 \\
\hline
\multirow{4}{*}{Scen.-III}
 & P-T & \underline{3.18} & 0.79 & \textbf{0.63} & 0.67 & 0.67 \\
 & P-S & \textbf{4.17} & \underline{4.96} & 4.68 & 4.48 & 4.44 \\
 & R-T & \underline{3.15} & 1.94 & 1.91 & 1.66 & \textbf{1.63} \\
 & R-S & \underline{3.51} & 1.50 & 1.28 & 1.21 & \textbf{1.20} \\
\hline
\multirow{8}{*}{Scen.-IV}
 & Seg1, P-T & \underline{1.97} & 1.39 & \textbf{1.35} & \textbf{1.35} & \textbf{1.35} \\
 & Seg1, P-S & \underline{4.09} & 3.45 & \textbf{3.40} & 3.42 & \textbf{3.40} \\
 & Seg1, R-T & \underline{5.07} & 3.60 & 3.16 & 3.35 & \textbf{2.61} \\
 & Seg1, R-S & \underline{5.95} & 4.23 & \textbf{3.81} & 4.10 & 3.92 \\
 & Seg2, P-T & 2.10 & 2.08 & \underline{2.12} & \textbf{2.04} & 2.08 \\
 & Seg2, P-S & \textbf{3.35} & 3.38 & \underline{3.44} & 3.40 & 3.41 \\
 & Seg2, R-T & \underline{3.75} & 3.36 & 3.15 & 3.35 & \textbf{2.86} \\
 & Seg2, R-S & \underline{2.61} & 2.39 & 2.36 & 2.35 & \textbf{2.18} \\
\hline
\end{tabular}
\label{tab:exp_error}
\vspace{1ex}
\footnotesize
\textbf{Note:} 
P = Position, 
T = Tip, 
R = Rotation, 
S = Shape. 
Errors are measured in $[\%]$ for position and $[^\circ/100$ mm] for rotation. 
\textbf{Bold} indicates the smallest (best) value in each row, and \underline{underline} indicates the largest (worst) value.
\end{table}

\section{Conclusion}
In this paper, we proposed a stochastic observer-based shape estimation framework for continuum robots within a polynomial curvature shape state space. 
The contribution lies in treating dynamic shape estimation as a multi-model problem in curvature space and applying an IMM–EKF over the polynomial curvature kinematics (PCK) family to adapt across regimes.
We target limited-sensing deployments; in hardware, a single tip pose measurement was used, while the framework supports general pose/position/orientation inputs.
Two IMM–EKF variants (SAIMM, DAIMM) were developed for dynamic configurations, and a noise-weighted observability matrix was introduced to guide sensor placement under sparse measurements.
\par
We evaluated the approach in simulations and hardware.
Baselines included the classical CC EKF and single-PCK EKFs using the same polynomial orders as the IMM candidates, so that improvements are attributable to multi-model adaptation rather than a change of parameterization.
Across free-bending, tip-load, distributed-load, and two-segment experiments, IMM–PCK reduced or matched errors relative to single-PCK EKFs and consistently outperformed the CC baseline under the same single-tip measurement; SAIMM showed the most uniform robustness across regimes, while DAIMM responded more aggressively to abrupt changes.
\par
\textit{Limitations and future work.} A current limitation is the omission of torsion in the PCK.
We will extend the PCK and observer to full $SE(3)$ (curvature and twist) for general 3D shape estimation under limited/sparse sensing.
Additional directions include integrating advanced control, exploring alternative filters (e.g., UKF), and combining basic static models for force/moment estimation.

\section*{Supplementary Appendix}
An extended Supplementary Appendix (Appendices A–C) is available as an ancillary file with the arXiv version of this article\cite{zhang2022stochastic}.

\section*{Multimedia Materials}
A supplementary video demonstrates the estimator’s performance in Scenarios~I–IV.
\section*{Acknowledgement}
The authors thank Isara Cholaseuk for implementing the actuator design in CAD, fabricating the 3D-printed components, and for helpful discussions during prototyping.
\bibliographystyle{IEEEtran}
\bibliography{bib/IEEEabrv,bib/general_continuum_robots,bib/constant_curvature,bib/cosserat_rod,bib/psuedo_rigid_body,bib/modal_approach,bib/polynomial_curve,bib/shape_sensing, bib/imm_ekf}

@article{webster2010design,
  title={Design and kinematic modeling of constant curvature continuum robots: A review},
  author={Webster III, Robert J and Jones, Bryan A},
  journal={The International Journal of Robotics Research},
  volume={29},
  number={13},
  pages={1661--1683},
  year={2010},
  publisher={SAGE Publications Sage UK: London, England}
}

@inproceedings{simaan2004dexterous,
  title={A dexterous system for laryngeal surgery},
  author={Simaan, Nabil and Taylor, Russell and Flint, Paul},
  booktitle={IEEE International Conference on Robotics and Automation, 2004. Proceedings. ICRA'04. 2004},
  volume={1},
  pages={351--357},
  year={2004},
  organization={IEEE}
}

@article{camarillo2008mechanics,
  title={Mechanics modeling of tendon-driven continuum manipulators},
  author={Camarillo, David B and Milne, Christopher F and Carlson, Christopher R and Zinn, Michael R and Salisbury, J Kenneth},
  journal={IEEE transactions on robotics},
  volume={24},
  number={6},
  pages={1262--1273},
  year={2008},
  publisher={IEEE}
}

@article{hannan2003kinematics,
  title={Kinematics and the implementation of an elephant's trunk manipulator and other continuum style robots},
  author={Hannan, Michael W and Walker, Ian D},
  journal={Journal of robotic systems},
  volume={20},
  number={2},
  pages={45--63},
  year={2003},
  publisher={Wiley Online Library}
}

@article{zhao2022modular,
  title={A Modular Continuum Manipulator for Aerial Manipulation and Perching},
  author={Zhao, Qianwen and Zhang, Guoqing and Jafarnejadsani, Hamidreza and Wang, Long},
  journal={arXiv preprint arXiv:2206.06246},
  year={2022}
}

@inproceedings{sears2006steerable,
address = {Beijing, China},
author = {Sears, Patrick and Dupont, Pierre},
booktitle = {2006 IEEE/RSJ International Conference on Intelligent Robots and Systems},
doi = {10.1109/IROS.2006.282072},
isbn = {1-4244-0258-1},
month = {oct},
pages = {2850--2856},
publisher = {IEEE},
title = {{A Steerable Needle Technology Using Curved Concentric Tubes}},
year = {2006}
}

@article{sitler2022modular,
month = {dec},
number = {6},
pages = {1--10},
title = {{A Modular Open-Source Continuum Manipulator for Underwater Remotely Operated Vehicles}},
volume = {14},
year = {2022}
}

@inproceedings{borgstadt2015multi,
  title={Multi-modal localization algorithm for catheter interventions},
  author={Borgstadt, Justin A and Zinn, Michael R and Ferrier, Nicola J},
  booktitle={2015 IEEE international conference on robotics and automation (ICRA)},
  pages={5350--5357},
  year={2015},
  organization={IEEE}
}

@inproceedings{chen2019model,
  title={Model-based estimation of the gravity-loaded shape and scene depth for a slim 3-actuator continuum robot with monocular visual feedback},
  author={Chen, Yuyang and Zeng, Lingyun and Zhu, Xiangyang and Xu, Kai and others},
  booktitle={2019 International Conference on Robotics and Automation (ICRA)},
  pages={4416--4421},
  year={2019},
  organization={IEEE}
}

@inproceedings{ataka2016real,
  title={Real-time pose estimation and obstacle avoidance for multi-segment continuum manipulator in dynamic environments},
  author={Ataka, Ahmad and Qi, Peng and Shiva, Ali and Shafti, Ali and Wurdemann, Helge and Liu, Hongbin and Althoefer, Kaspar},
  booktitle={2016 IEEE/RSJ International Conference on Intelligent Robots and Systems (IROS)},
  pages={2827--2832},
  year={2016},
  organization={IEEE}
}

@inproceedings{loo2019non,
  title={Non-linear system identification and state estimation in a pneumatic based soft continuum robot},
  author={Loo, Junn Yong and Kong, Kah Chun and Tan, Chee Pin and Nurzaman, Surya Girinatha},
  booktitle={2019 IEEE Conference on control technology and applications (CCTA)},
  pages={39--46},
  year={2019},
  organization={IEEE}
}

@article{peng2024tendon,
  title={A Tendon-Driven Continuum Manipulator With Robust Shape Estimation by Multiple IMUs},
  author={Peng, Rui and Wang, Yu and Lu, Peng},
  journal={IEEE Robotics and Automation Letters},
  year={2024},
  publisher={IEEE}
}

@article{santoso2021origami,
  title={An origami continuum robot capable of precise motion through torsionally stiff body and smooth inverse kinematics},
  author={Santoso, Junius and Onal, Cagdas D},
  journal={Soft Robotics},
  volume={8},
  number={4},
  pages={371--386},
  year={2021},
  publisher={Mary Ann Liebert, Inc., publishers 140 Huguenot Street, 3rd Floor New~…}
}

@article{sun2023enhancing,
  title={Enhancing torsional stiffness of continuum robots using 3-D topology optimized flexure joints},
  author={Sun, Yilun and Lueth, Tim C},
  journal={IEEE/ASME Transactions on Mechatronics},
  volume={28},
  number={4},
  pages={1844--1852},
  year={2023},
  publisher={IEEE}
}

@article{gunderman2023non,
  title={Non-metallic MR-guided concentric tube robot for intracerebral hemorrhage evacuation},
  author={Gunderman, Anthony L and Sengupta, Saikat and Siampli, Eleni and Sigounas, Dimitri and Kellner, Christopher and Oluigbo, Chima and Sharma, Karun and Godage, Isuru and Cleary, Kevin and Chen, Yue},
  journal={IEEE Transactions on Biomedical Engineering},
  volume={70},
  number={10},
  pages={2895--2904},
  year={2023},
  publisher={IEEE}
}

@article{qiu2025actuator,
  title={An actuator space optimal kinematic path tracking framework for tendon-driven continuum robots: Theory, algorithm and validation},
  author={Qiu, Ke and Zhang, Hongye and Zhang, Jingyu and Xiong, Rong and Lu, Haojian and Wang, Yue},
  journal={The International Journal of Robotics Research},
  volume={44},
  number={6},
  pages={1006--1034},
  year={2025},
  publisher={SAGE Publications Sage UK: London, England}
}

@article{rogatinsky2023multifunctional,
  title={A multifunctional soft robot for cardiac interventions},
  author={Rogatinsky, Jacob and Recco, Dominic and Feichtmeier, Joseph and Kang, Yuchen and Kneier, Nicholas and Hammer, Peter and O’Leary, Edward and Mah, Douglas and Hoganson, David and Vasilyev, Nikolay V and others},
  journal={Science Advances},
  volume={9},
  number={43},
  pages={eadi5559},
  year={2023},
  publisher={American Association for the Advancement of Science}
}

@article{russo2023continuum,
  title={Continuum robots: An overview},
  author={Russo, Matteo and Sadati, Seyed Mohammad Hadi and Dong, Xin and Mohammad, Abdelkhalick and Walker, Ian D and Bergeles, Christos and Xu, Kai and Axinte, Dragos A},
  journal={Advanced Intelligent Systems},
  volume={5},
  number={5},
  pages={2200367},
  year={2023},
  publisher={Wiley Online Library}
}

@inproceedings{jones2009three,
  title={Three dimensional statics for continuum robotics},
  author={Jones, Bryan A and Gray, Ricky L and Turlapati, Krishna},
  booktitle={2009 IEEE/RSJ International Conference on Intelligent Robots and Systems},
  pages={2659--2664},
  year={2009},
  organization={IEEE}
}

@article{rucker2011statics,
  title={Statics and dynamics of continuum robots with general tendon routing and external loading},
  author={Rucker, D Caleb and Webster III, Robert J},
  journal={IEEE Transactions on Robotics},
  volume={27},
  number={6},
  pages={1033--1044},
  year={2011},
  publisher={IEEE}
}

@article{till2019real,
  title={Real-time dynamics of soft and continuum robots based on Cosserat rod models},
  author={Till, John and Aloi, Vincent and Rucker, Caleb},
  journal={The International Journal of Robotics Research},
  volume={38},
  number={6},
  pages={723--746},
  year={2019},
  publisher={SAGE Publications Sage UK: London, England}
}

@article{boyer2020dynamics,
  title={Dynamics of continuum and soft robots: A strain parameterization based approach},
  author={Boyer, Frederic and Lebastard, Vincent and Candelier, Fabien and Renda, Federico},
  journal={IEEE Transactions on Robotics},
  volume={37},
  number={3},
  pages={847--863},
  year={2020},
  publisher={IEEE}
}

@article{renda2018discrete,
author = {Renda, Federico and Boyer, Frederic and Dias, Jorge and Seneviratne, Lakmal},
doi = {10.1109/TRO.2018.2868815},
issn = {1552-3098},
journal = {IEEE Transactions on Robotics},
month = {dec},
number = {6},
pages = {1518--1533},
title = {{Discrete Cosserat Approach for Multisection Soft Manipulator Dynamics}},
volume = {34},
year = {2018}
}

@inproceedings{mahoney2016inseparable,
  title={On the inseparable nature of sensor selection, sensor placement, and state estimation for continuum robots or “where to put your sensors and how to use them”},
  author={Mahoney, Arthur W and Bruns, Trevor L and Swaney, Philip J and Webster, Robert J},
  booktitle={2016 IEEE International Conference on Robotics and Automation (ICRA)},
  pages={4472--4478},
  year={2016},
  organization={IEEE}
}

@article{anderson2017continuum,
  title={Continuum reconfigurable parallel robots for surgery: Shape sensing and state estimation with uncertainty},
  author={Anderson, Patrick L and Mahoney, Arthur W and Webster, Robert James},
  journal={IEEE robotics and automation letters},
  volume={2},
  number={3},
  pages={1617--1624},
  year={2017},
  publisher={IEEE}
}

@article{lilge2022continuum,
  title={Continuum robot state estimation using Gaussian process regression on SE (3)},
  author={Lilge, Sven and Barfoot, Timothy D and Burgner-Kahrs, Jessica},
  journal={The International Journal of Robotics Research},
  volume={41},
  number={13-14},
  pages={1099--1120},
  year={2022},
  publisher={SAGE Publications Sage UK: London, England}
}

@article{ferguson2024unified,
  title={Unified shape and external load state estimation for continuum robots},
  author={Ferguson, James M and Rucker, D Caleb and Webster, Robert J},
  journal={IEEE Transactions on Robotics},
  year={2024},
  publisher={IEEE}
}

@article{hirose1993biologically,
  title={Biologically inspired robots},
  author={Hirose, Shigeo},
  journal={Snake-Like Locomotors and Manipulators},
  year={1993},
  publisher={Oxford Science Publications}
}

@article{dupont2022continuum,
  title={Continuum robots for medical interventions},
  author={Dupont, Pierre E and Simaan, Nabil and Choset, Howie and Rucker, Caleb},
  journal={Proceedings of the IEEE},
  volume={110},
  number={7},
  pages={847--870},
  year={2022},
  publisher={IEEE}
}

@article{simaan2018medical,
  title={Medical technologies and challenges of robot-assisted minimally invasive intervention and diagnostics},
  author={Simaan, Nabil and Yasin, Rashid M and Wang, Long},
  journal={Annual Review of Control, Robotics, and Autonomous Systems},
  volume={1},
  pages={465--490},
  year={2018},
  publisher={Annual Reviews}
}

@article{rao2021model,
  title={How to model tendon-driven continuum robots and benchmark modelling performance},
  author={Rao, Priyanka and Peyron, Quentin and Lilge, Sven and Burgner-Kahrs, Jessica},
  journal={Frontiers in Robotics and AI},
  volume={7},
  pages={630245},
  year={2021},
  publisher={Frontiers Media SA}
}

@article{rone2013continuum,
  title={Continuum robot dynamics utilizing the principle of virtual power},
  author={Rone, William S and Ben-Tzvi, Pinhas},
  journal={IEEE Transactions on Robotics},
  volume={30},
  number={1},
  pages={275--287},
  year={2013},
  publisher={IEEE}
}

@article{peng2025dexterous,
  title={A dexterous and compliant aerial continuum manipulator for cluttered and constrained environments},
  author={Peng, Rui and Wang, Yu and Lu, Minghao and Lu, Peng},
  journal={Nature Communications},
  volume={16},
  number={1},
  pages={889},
  year={2025},
  publisher={Nature Publishing Group UK London}
}

@article{yuan2012multiple,
  title={A multiple IMM estimation approach with unbiased mixing for thrusting projectiles},
  author={Yuan, Ting and Bar-Shalom, Yaakov and Willett, Peter and Mozeson, E and Pollak, S and Hardiman, David},
  journal={IEEE Transactions on Aerospace and Electronic Systems},
  volume={48},
  number={4},
  pages={3250--3267},
  year={2012},
  publisher={IEEE}
}

@article{xie2019adaptive,
  title={Adaptive transition probability matrix-based parallel IMM algorithm},
  author={Xie, Guo and Sun, Lanlan and Wen, Tao and Hei, Xinhong and Qian, Fucai},
  journal={IEEE Transactions on Systems, Man, and Cybernetics: Systems},
  volume={51},
  number={5},
  pages={2980--2989},
  year={2019},
  publisher={IEEE}
}

@article{lee2023improved,
  title={An Improved Interacting Multiple Model Algorithm With Adaptive Transition Probability Matrix Based on the Situation},
  author={Lee, In Ho and Park, Chan Gook},
  journal={International Journal of Control, Automation and Systems},
  volume={21},
  number={10},
  pages={3299--3312},
  year={2023},
  publisher={Springer}
}

@article{hermann1977nonlinear,
  title={Nonlinear controllability and observability},
  author={Hermann, Robert and Krener, Arthur},
  journal={IEEE Transactions on automatic control},
  volume={22},
  number={5},
  pages={728--740},
  year={1977},
  publisher={IEEE}
}

@article{
    doi:10.1126/scirobotics.abm6074,
    author = {Steven Macenski  and Tully Foote  and Brian Gerkey  and Chris Lalancette  and William Woodall },
    title = {Robot Operating System 2: Design, architecture, and uses in the wild},
    journal = {Science Robotics},
    volume = {7},
    number = {66},
    pages = {eabm6074},
    year = {2022},
    doi = {10.1126/scirobotics.abm6074},
    URL = {https://www.science.org/doi/abs/10.1126/scirobotics.abm6074}
}

@article{mazor1998interacting,
  title={Interacting multiple model methods in target tracking: a survey},
  author={Mazor, Efim and Averbuch, Amir and Bar-Shalom, Yakov and Dayan, Joshua},
  journal={IEEE Transactions on aerospace and electronic systems},
  volume={34},
  number={1},
  pages={103--123},
  year={1998},
  publisher={IEEE}
}

@article{dingler2022state,
  title={State estimation with the Interacting Multiple Model (IMM) method},
  author={Dingler, Sebastian},
  journal={arXiv preprint arXiv:2207.04875},
  year={2022}
}

@article{zhang2022stochastic,
  title={Stochastic Adaptive Estimation in Polynomial Curvature Shape State Space for Continuum Robots},
  author={Zhang, Guoqing and Wang, Long},
  journal={arXiv preprint arXiv:2210.08427},
  year={2022}
}

@article{della2019control,
   author={Santina, Cosimo Della and Rus, Daniela},
  journal={IEEE Robotics and Automation Letters}, 
  title={Control Oriented Modeling of Soft Robots: The Polynomial Curvature Case}, 
  year={2020},
  volume={5},
  number={2},
  pages={290-298},
  doi={10.1109/LRA.2019.2955936}
}

@article{chirikjian1994modal,
  title={A modal approach to hyper-redundant manipulator kinematics},
  author={Chirikjian, Gregory S and Burdick, Joel W},
  journal={IEEE Transactions on Robotics and Automation},
  volume={10},
  number={3},
  pages={343--354},
  year={1994},
  publisher={IEEE}
}

@article{wang2019geometric,
  title={Geometric calibration of continuum robots: Joint space and equilibrium shape deviations},
  author={Wang, Long and Simaan, Nabil},
  journal={IEEE Transactions on Robotics},
  volume={35},
  number={2},
  pages={387--402},
  year={2019},
  publisher={IEEE}
}

@article{sadati2023reduced,
  title={Reduced order modeling and model order reduction for continuum manipulators: an overview},
  author={Sadati, SM Hadi and Naghibi, S Elnaz and Da Cruz, Lyndon and Bergeles, Christos},
  journal={Frontiers in Robotics and AI},
  volume={10},
  pages={1094114},
  year={2023},
  publisher={Frontiers Media SA}
}

@article{shi2016shape,
author = {Shi, Chaoyang and Luo, Xiongbiao and Qi, Peng and Li, Tianliang and Song, Shuang and Najdovski, Zoran and Fukuda, Toshio and Ren, Hongliang},
doi = {10.1109/TBME.2016.2622361},
issn = {15582531},
journal = {IEEE Transactions on Biomedical Engineering},
number = {8},
pages = {1665--1678},
pmid = {27810796},
title = {{Shape sensing techniques for continuum robots in minimally invasive surgery: A survey}},
volume = {64},
year = {2017}
}

@inproceedings{cheng2022orientation,
  title={Orientation to Pose: Continuum Robots Shape Reconstruction Based on the Multi-Attitude Solving Approach},
  author={Cheng, Hao and Xu, Hejie and Shang, Hongji and Wang, Xueqian and Liu, Houde and Liang, Bin},
  booktitle={2022 International Conference on Robotics and Automation (ICRA)},
  pages={3203--3209},
  year={2022},
  organization={IEEE}
}

@inproceedings{kim2014optimizing,
  title={Optimizing curvature sensor placement for fast, accurate shape sensing of continuum robots},
  author={Kim, Beobkyoon and Ha, Junhyoung and Park, Frank C and Dupont, Pierre E},
  booktitle={2014 IEEE international conference on robotics and automation (ICRA)},
  pages={5374--5379},
  year={2014},
  organization={IEEE}
}

@article{song2015electromagnetic,
  title={Electromagnetic positioning for tip tracking and shape sensing of flexible robots},
  author={Song, Shuang and Li, Zheng and Yu, Haoyong and Ren, Hongliang},
  journal={IEEE Sensors Journal},
  volume={15},
  number={8},
  pages={4565--4575},
  year={2015},
  publisher={IEEE}
}

@article{rao2022shape,
  title={Shape representation and modeling of tendon-driven continuum robots using euler arc splines},
  author={Rao, Priyanka and Peyron, Quentin and Burgner-Kahrs, Jessica},
  journal={IEEE Robotics and Automation Letters},
  volume={7},
  number={3},
  pages={8114--8121},
  year={2022},
  publisher={IEEE}
}

@article{orekhov2023lie,
  title={Lie group formulation and sensitivity analysis for shape sensing of variable curvature continuum robots with general string encoder routing},
  author={Orekhov, Andrew L and Ahronovich, Elan Z and Simaan, Nabil},
  journal={IEEE Transactions on Robotics},
  year={2023},
  publisher={IEEE}
}

@inproceedings{gandhi2023shape,
  title={Shape Control of Variable Length Continuum Robots using Clothoid-based Visual Servoing},
  author={Gandhi, Abhinav and Chiang, Shou-Shan and Onal, Cagdas D and Calli, Berk},
  booktitle={2023 IEEE/RSJ International Conference on Intelligent Robots and Systems (IROS)},
  pages={6379--6386},
  year={2023},
  organization={IEEE}
}

@article{roesthuis2016steering,
  title={Steering of multisegment continuum manipulators using rigid-link modeling and FBG-based shape sensing},
  author={Roesthuis, Roy J and Misra, Sarthak},
  journal={IEEE transactions on robotics},
  volume={32},
  number={2},
  pages={372--382},
  year={2016},
  publisher={IEEE}
}

@article{su2009pseudorigid,
    author = {Su, Hai-Jun},
    title = "{A Pseudorigid-Body 3R Model for Determining Large Deflection of Cantilever Beams Subject to Tip Loads}",
    journal = {Journal of Mechanisms and Robotics},
    volume = {1},
    number = {2},
    year = {2009},
    month = {01},
    issn = {1942-4302},
    doi = {10.1115/1.3046148},
    note = {021008},
}

@inproceedings{khoshnam2015robotics,
  title={A robotics-assisted catheter manipulation system for cardiac ablation with real-time force estimation},
  author={Khoshnam, Mahta and Khalaji, Iman and Patel, Rajni V},
  booktitle={2015 IEEE/RSJ International Conference on Intelligent Robots and Systems (IROS)},
  pages={3202--3207},
  year={2015},
  organization={IEEE}
}

@inproceedings{huang20193d,
  title={A 3D static modeling method and experimental verification of continuum robots based on pseudo-rigid body theory},
  author={Huang, Shaoping and Meng, Deshan and Wang, Xueqian and Liang, Bin and Lu, Weining},
  booktitle={2019 IEEE/RSJ International Conference on Intelligent Robots and Systems (IROS)},
  pages={4672--4677},
  year={2019},
  organization={IEEE}
}

@article{venkiteswaran2019shape,
  title={Shape and contact force estimation of continuum manipulators using pseudo rigid body models},
  author={Venkiteswaran, Venkatasubramanian Kalpathy and Sikorski, Jakub and Misra, Sarthak},
  journal={Mechanism and machine theory},
  volume={139},
  pages={34--45},
  year={2019},
  publisher={Elsevier}
}

@article{qiao2021force,
  title={Force from shape—estimating the location and magnitude of the external force on flexible instruments},
  author={Qiao, Qiao and Borghesan, Gianni and De Schutter, Joris and Vander Poorten, Emmanuel},
  journal={IEEE Transactions on Robotics},
  volume={37},
  number={5},
  pages={1826--1833},
  year={2021},
  publisher={IEEE}
}

@inproceedings{nahvi1996noise,
  title={The noise amplification index for optimal pose selection in robot calibration},
  author={Nahvi, Ali and Hollerbach, John M},
  booktitle={Proceedings of IEEE international conference on robotics and automation},
  volume={1},
  pages={647--654},
  year={1996},
  organization={IEEE}
}

@inproceedings{qiao2019estimating,
  title={Estimating and localizing external forces applied on flexible instruments by shape sensing},
  author={Qiao, Qiao and Willems, Dries and Borghesan, Gianni and Ourak, Mouloud and De Schutter, Joris and Vander Poorten, Emmanuel},
  booktitle={2019 19th International Conference on Advanced Robotics (ICAR)},
  pages={227--233},
  year={2019},
  organization={IEEE}
}

@article{al2021fbg,
  title={Fbg-based estimation of external forces along flexible instrument bodies},
  author={Al-Ahmad, Omar and Ourak, Mouloud and Vlekken, Johan and Vander Poorten, Emmanuel},
  journal={Frontiers in Robotics and AI},
  volume={8},
  pages={718033},
  year={2021},
  publisher={Frontiers Media SA}
}

@inproceedings{sefati2017highly,
  title={A highly sensitive fiber Bragg grating shape sensor for continuum manipulators with large deflections},
  author={Sefati, Shahriar and Pozin, Michael and Alambeigi, Farshid and Iordachita, Iulian and Taylor, Russell H and Armand, Mehran},
  booktitle={2017 IEEE SENSORS},
  pages={1--3},
  year={2017},
  organization={IEEE}
}

@inproceedings{rao2021using,
  title={Using Euler curves to model continuum robots},
  author={Rao, Priyanka and Peyron, Quentin and Burgner-Kahrs, Jessica},
  booktitle={2021 IEEE International Conference on Robotics and Automation (ICRA)},
  pages={1402--1408},
  year={2021},
  organization={IEEE}
}
\balance
\begin{IEEEbiography}[{\includegraphics[width=1in,height=1.25in,clip,keepaspectratio]{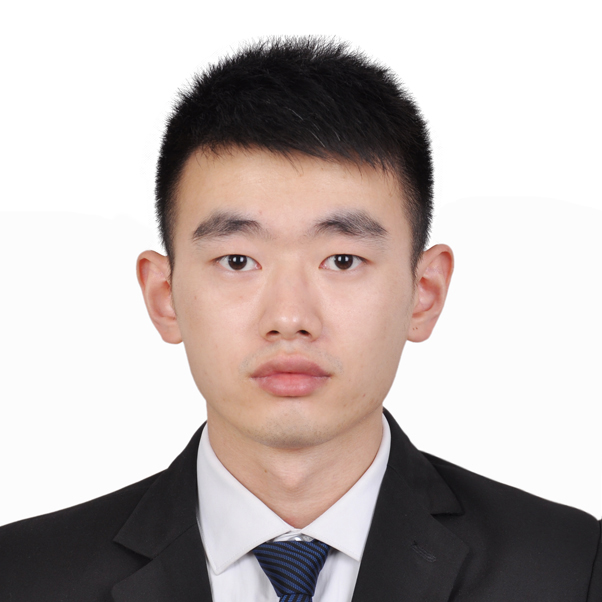}}]{Guoqing Zhang} received the B.E. degree in processing equipment and control engineering from Tianjin University, Tianjin, China, in 2014 and the M.E. degree in Chemical machinery from Xi'an Jiaotong University, Xi'an, China and the M.S. degree in mechanical engineering from Columbia University, New York, NY, USA, in 2019. He is currently a Ph.D. student studying surgical continuum robotics at Stevens Institute of Technology, Hoboken, NJ USA. His research interests include modeling, control, and estimation theory of continuum manipulators and other robotic systems. \end{IEEEbiography}

\begin{IEEEbiography}[{\includegraphics[width=1in,height=1.25in,clip,keepaspectratio]{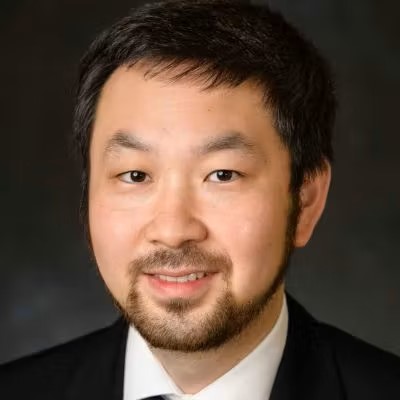}}]{Long Wang}  received the B.S. degree from Tsinghua University, Beijing, China, in 2010, the M.S. degree from Columbia University,
New York, NY, USA, in 2012, and the Ph.D. degree from Vanderbilt University, Nashville, TN, USA, in 2019, all in mechanical engineering.

He was a Postdoctoral Researcher with the Department of Mechanical Engineering, Columbia University, New York, NY, from 2018 to 2019. He is currently an Assistant Professor with the Department of Mechanical Engineering, Stevens Institute of Technology, Hoboken, NJ, USA. His research interests include modeling and calibration of continuum robots, surgical robotics, telemanipulation, manipulation, and grasping.
\end{IEEEbiography}

\end{document}